\PassOptionsToPackage{table,xcdraw}{xcolor}
\documentclass[a4paper,fleqn]{cas-dc}

\renewcommand{\arraystretch}{1.5}

\newcommand{\cs}{\textsc{ClickSight}\xspace}
\newcommand{\compl}{\textit{Completeness}\xspace}
\newcommand{\corr}{\textit{Correctness}\xspace}
\newcommand{\just}{\textit{Justifiedness}\xspace}
\newcommand{\compr}{\textit{Comprehensibility}\xspace}

\newcommand{\zs}{\textit{Zero-shot}\xspace}
\newcommand{\ct}{\textit{Chain-of-Thought}\xspace}
\newcommand{\cop}{\textit{Chain-of-Prompts}\xspace}
\newcommand{\ps}{PharmaSim\xspace}
\newcommand{\bl}{Beer's Law Lab\xspace}

\definecolor{envframe}{RGB}{0,32,96}  %
\definecolor{lightpink}{RGB}{255,230,235}
\definecolor{lightgreen}{RGB}{227,243,213}
\definecolor{lightblue}{RGB}{214,235,242}
\definecolor{lightsalmon}{RGB}{255,232,213}

\definecolor{suppcol}{RGB}{0,90,181}
\definecolor{unsuppcol}{RGB}{201,90,0}

\IfFileExists{stix.sty}{}{%
  \IfFileExists{lmodern.sty}{\usepackage{lmodern}}%
    {\usepackage{mathptmx}\usepackage[scaled=.92]{helvet}\usepackage{courier}}%
}

\usepackage{graphicx}
\usepackage{subcaption}
\usepackage{siunitx}
\usepackage{multirow}
\usepackage{amsmath,amssymb,amsfonts}
\usepackage{amsthm}
\usepackage{booktabs}
\usepackage{adjustbox}
\usepackage{makecell}
\usepackage{mathrsfs}
\input{ursfs.fd}
\DeclareFontShape{U}{rsfs}{m}{n}{<-6>rsfs5<6-8.5>rsfs7<8.5->rsfs10}{}
\begingroup\fontencoding{T1}\fontfamily{cmss}\selectfont\endgroup %
\DeclareFontShape{T1}{cmss}{bx}{sc}{<->ssub*cmss/bx/n}{}
\usepackage[title]{appendix}
\usepackage{textcomp}
\usepackage{manyfoot}
\usepackage{svg}
\usepackage{cleveref}
\usepackage{algorithm}
\usepackage{algorithmicx}
\usepackage{algpseudocode}
\usepackage{listings}
\usepackage{tcolorbox}
\tcbuselibrary{raster}
\usepackage{xspace}
\usepackage{tabularx}
\usepackage{float}
\usepackage{placeins} %
\usepackage{threeparttable}
\usepackage{enumitem}
\usepackage{ragged2e}
\usepackage{balance}
\usepackage{microtype}   %
\usepackage{xurl}        %
\hypersetup{
    colorlinks=true,
    linkcolor=blue!50!black,
    citecolor=teal!50!black,
    urlcolor=blue!40!black
}

\DeclareFontFamily{OT1}{cmtt}{}
\DeclareFontShape{OT1}{cmtt}{bx}{n}{<->cmttb10}{}
\DeclareFontShape{OT1}{cmtt}{b}{n}{<->cmttb10}{}
\tcbuselibrary{breakable}

\renewcommand{\arraystretch}{1.3}

\usepackage[authoryear]{natbib}
\bibpunct{(}{)}{;}{a}{,}{,}

\usepackage{lastpage}
\AtBeginDocument{\renewcommand{\lastpage}{\pageref{LastMainPage}}}
\def\tsc#1{\csdef{#1}{\textsc{\lowercase{#1}}\xspace}}
\tsc{WGM}
\tsc{QE}

\ExplSyntaxOn
\cs_set:Npn \__first_footerline:
  { \group_begin: \small \sffamily \__short_authors: :~ { \rmfamily \itshape Preprint } \group_end: }
\ExplSyntaxOff
\begin{document}

\shorttitle{ClickSight: Interpreting Student Clickstreams Through Learning Strategies}
\shortauthors{Radmehr et al.}

\title[mode = title]{ClickSight: An LLM Framework for Interpreting Student Clickstreams Through Learning Strategies}

\author[1]{Bahar Radmehr}[orcid=0009-0008-8034-4853]
\cormark[1]
\ead{bahar.radmehr@epfl.ch}
\affiliation[1]{organization={Machine Learning for Education Laboratory (ML4ED), School of Computer and Communication Sciences, EPFL},
            addressline={INF Building, Station 14},
            city={Lausanne},
            postcode={1015},
            country={Switzerland}}

\author[1]{Lan Tran}[orcid=0009-0003-1404-3263]
\ead{lan.tranhuong@epfl.ch}

\author[1]{Ekaterina Shved}[orcid=0009-0004-8915-3640]
\ead{kate.kutsenok@epfl.ch}

\author[1]{Fatma Bet\"ul G\"ure\c{s}}[orcid=0000-0002-8664-1022]
\ead{fatma-betul.gures@epfl.ch}

\author[2]{Adish Singla}[orcid=0000-0001-9922-0668]
\ead{adishs@mpi-sws.org}
\affiliation[2]{organization={Max Planck Institute for Software Systems},
            addressline={Campus E1 5},
            city={Saarbr\"ucken},
            postcode={66123},
            country={Germany}}

\author[1]{Tanja K\"aser}[orcid=0000-0003-0672-0415]
\ead{tanja.kaeser@epfl.ch}

\cortext[1]{Corresponding author}

\begin{abstract}
Digital learning environments produce rich clickstream data that can reveal how students approach learning tasks, yet interpreting these fine-grained traces often requires substantial expert effort. We introduce \cs, a framework for interpreting student clickstreams through learning strategies defined in the literature. Given a student clickstream and a predefined set of learning strategies, \cs uses a staged LLM-based pipeline to identify strategy-relevant evidence, assess adherence to each strategy, and generate an evidence-grounded interpretation of the student's behavior. We evaluate \cs across two learning environments, 365 students, and 12 model backbones using an expert-grounded quality rubric and an LLM judge calibrated with human-expert ratings. Our results show that \cs can generate high-quality strategy-based interpretations across different models and environments, with reasoning models performing more consistently than instruction-tuned counterparts. We further find that providing external clickstream-extracted guidance can improve interpretation quality for strategies that require precise quantitative information from the trace. By connecting fine-grained clickstreams with theoretically defined learning strategies and preserving the evidence underlying each judgment, \cs provides a flexible approach for studying student behavior in digital learning environments.
\end{abstract}


\maketitle

\section{Introduction}\label{intro}
Digital learning environments are becoming increasingly widespread, offering opportunities for personalized, scalable, and data-driven educational experiences~\citep{rutten2012learning,moore2013interactive}. 
These environments capture detailed temporal records of students' actions, commonly referred to as clickstream data, which can provide valuable insights into students' behaviors, informing instructors and supporting the adaptation of instruction to students' needs~\citep{DBLP:conf/lak/ParkDRSW17}. However, translating raw clickstream logs into meaningful interpretations of students' learning behaviors remains challenging. Clickstream data are highly granular and structured around the specific features and interactions of individual learning environments, making their interpretation cognitively demanding, time-intensive, and dependent on specialized expertise~\citep{clickstream-difficulty}.

A large body of prior work has therefore focused on deriving meaningful representations of student behavior from clickstream data using machine learning techniques. Prior approaches include designing domain-specific features to capture strategy use~\citep{bll,DBLP:journals/aiedu/KaserS20,perez2018control,FRA17}, clustering student behaviors~\citep{bll,wang2016unsupervised,2025.EDM.cluster,Rohani2024ClickTree,cicchinelli2018finding}, and applying sequential pattern mining to uncover common interaction traces~\citep{DBLP:conf/lak/ParkDRSW17,yurum2023use,edm2020_classifier_pattern,aied2024_seq_dashboard,perez2017identifying,DBLP:journals/aiedu/ZhangBH22}. Other work has trained supervised models using expert-labeled data~\citep{edm2020_classifier_pattern,DBLP:conf/aied/GuresNRRK25,Li2022} or applied knowledge-tracing techniques to infer latent student states~\citep{aied2024knowledge_tracing,DBLP:journals/aiedu/KaserS20}. While these approaches have successfully extracted meaningful patterns from clickstream data, translating low-level interactions into interpretable learning behaviors typically requires substantial expert input, for example to define features, construct labels, specify relevant patterns, or interpret the resulting representations. This reliance on expert-driven, often domain-specific processes makes it difficult to transfer such approaches across learning environments and limits their scalability. 

Large Language Models (LLMs) have the potential to alleviate these limitations by reducing the need for such expert effort, given their capabilities for pattern recognition, abstraction, and reasoning~\citep{DBLP:journals/corr/abs-2402-01580,Mirchandani2023LargeLM}. In AI for Education, however, LLMs have primarily been applied to semantically rich textual data, supporting tasks such as automated grading, feedback generation, and qualitative coding~\citep{yan_practical_2024}. More recent work has begun applying LLMs to behavioral and non-textual educational data, but typically through fine-tuning, precomputed indicators, or engineered representations~\citep{liu_integrating_2025,2025.EDM.long-papers.170,davalos_llms_2025}. It therefore remains underexplored whether LLMs can directly interpret clickstream data to derive higher-level behavioral insights grounded in theoretically defined learning strategies.

To address this gap, we introduce \cs, a framework for interpreting student clickstreams using learning strategies defined in the literature. Inspired by how experts approach this task, \cs uses an in-context LLM-based \cop pipeline that first filters the clickstream to retrieve strategy-relevant evidence and analyzes what this evidence indicates about the student’s adherence to the strategy. It then assigns a strategy-adherence level with an explicit, evidence-grounded justification and combines the per-strategy assessments into a coherent interpretation of the student's behavior.

We extensively evaluate these interpretations across two learning environments from different domains: \ps~\citep{ps}, a virtual pharmacy for training diagnostic reasoning, and \bl~\citep{bll}, an interactive chemistry simulation for training inquiry skills. To assess the quality of the generated interpretations, we develop a four-criterion rubric grounded in human-centered explanation theory~\cite{rubric1} and refined with domain-expert input. We first conduct expert evaluations on a subset of the interpretations and then develop an LLM-based judge that achieves high agreement with human raters, enabling evaluation at scale. Our experiments address the following research questions:
\begin{enumerate}
\item Can a state-of-the-art LLM (GPT-5) generate high-quality interpretations of student clickstream data \textbf{(RQ1)}?
\item Do these findings generalize across models of different sizes and with different reasoning capabilities \textbf{(RQ2)}?
\item Can adding external information improve the quality of strategy-based interpretations \textbf{(RQ3)}?
\end{enumerate}

Our results show that \cs can generate high-quality strategy-based interpretations of student clickstreams using GPT-5. These findings mostly generalize across models, with interpretation quality depending more on reasoning capability than model size. Finally, while generic expert-generated guidance provides limited benefits, precise automated summaries of student actions substantially improve interpretation quality, particularly for more complex learning strategies. By linking low-level clickstream data to high-level learning strategies, \cs offers a scalable way to interpret student behavior with reduced expert effort.

We provide our implementation of the \cs pipeline in our GitHub repository\footnote{\label{github}\url{https://github.com/epfl-ml4ed/ClickSight-CEAI}}.

\section{Related Work}

Our work builds on two main lines of research. The first studies how interaction data can be used to understand student behavior and learning strategies. The second examines how LLMs can be used to analyze educational data, including behavioral and non-textual data. We provide an overview of both areas to position our use of LLMs for generating strategy-based interpretations directly from student clickstreams.

\subsection{Clickstream-Based Learning Analysis}

Clickstream data have been analyzed in several ways to understand how students engage with learning environments, including predicting learning outcomes, identifying behavioral patterns, and studying the strategies students use~\citep{DBLP:conf/lak/ParkDRSW17,wang2016unsupervised,Rohani2024ClickTree}. A large body of work has focused specifically on learning strategies, which describe goal-oriented behaviors through which students approach, regulate, and carry out learning tasks~\citep{zimmerman2002becoming}. These include domain-specific strategies, such as controlling variables during inquiry and systematic questioning during differential diagnosis~\citep{bll,10.1145/3448139.3448169,DBLP:conf/aied/GuresNRRK25,ganaiem2025towards}, as well as self-regulatory processes such as planning, monitoring, and reflection~\citep{DBLP:conf/lak/RakovicFGSKLMBM22,DBLP:conf/edm/NathGR23,DBLP:conf/lak/LiG24,DBLP:conf/lak/NathGFR24}. Studying these strategies from clickstreams provides a way to connect observable student actions to theoretically meaningful descriptions of how students approach learning.

To identify theoretically defined strategies from clickstreams, domain experts and learning-science researchers typically operationalize them in terms of observable actions or sequences. Some approaches use expert-defined features or rules to map student actions onto learning strategies or self-regulated learning processes, which can inform instructional decision-making~\citep{bll,DBLP:conf/aied/GuresNRRK25,Li2022,DBLP:conf/lak/ChengGLR0FJTGS25}. For example, \cite{lifloraengine2025} use a theory-informed trace parser to map fine-grained student actions onto self-regulated learning processes and then use the detected processes to provide adaptive scaffolding. Other approaches use human-coded trace segments to train supervised models that detect strategy-related behaviors such as planning, hypothesis testing, and controlled experimentation~\citep{PedroUMUI,Orlando,PedroBGMN13}. These methods make theoretically defined constructs measurable from interaction data, but require expert knowledge to determine how behavioral traces should be represented, mapped, or labeled.

A complementary line of work identifies behavioral structure from the traces rather than beginning with a predefined strategy. Sequential pattern mining and temporal analysis have been used to uncover recurring ways in which students interact with learning environments~\citep{DBLP:conf/lak/ParkDRSW17,yurum2023use,edm2020_classifier_pattern,DBLP:conf/edm/GabbayC23,DBLP:conf/lak/WangLDHS24}. Clustering approaches similarly group students based on similarities in their interaction patterns or strategy use, producing behavioral profiles that can be compared in terms of learning approaches and outcomes~\citep{bll,wang2016unsupervised,2025.EDM.cluster,Rohani2024ClickTree,cicchinelli2018finding,DBLP:conf/lak/ZhangTC22,DBLP:conf/lak/SongLXLL24,DBLP:conf/lak/KimLJX25}. Such methods reduce the need to specify particular behavioral profiles in advance, but experts are still needed to interpret the discovered patterns or groups and determine their educational meaning.

Together, these approaches have enabled clickstream data to support increasingly detailed analyses of student behavior and learning strategies. However, deriving educationally meaningful interpretations still requires substantial expert involvement. For theory-driven analyses, experts must determine how abstract strategies map onto observable actions, features, rules, or annotations. For data-driven analyses, experts must interpret the behavioral patterns that emerge. These dependencies make analyses difficult to transfer across strategies and learning environments~\citep{bodily_review_2017,DBLP:conf/lak/Osakwe24,DBLP:conf/lak/Song0BKT25,valletorre2024sequence}.

\subsection{LLMs for Educational Data Analysis}

LLMs offer new possibilities for automating educational data analysis, but most existing applications operate on semantically rich textual inputs~\citep{yan_practical_2024}. They have been used to grade and provide feedback on student answers, essays, and programming assignments, often using rubrics or other task-specific guidance~\citep{chu_enhancing_2025,chu_llm-powered_2025,frohn_automated_2025,sesler_can_2025,xiao_human-ai_2025,cipriano_leveraging_2025,barros_large_2025}. These applications demonstrate that LLMs can evaluate complex student work, but the inputs themselves already express much of the information being interpreted in natural language or other semantically rich forms.

More closely related to our setting, LLMs have been used for qualitative coding, where observations are interpreted according to predefined theoretical or empirical categories. Studies have applied LLMs to identify self-regulated learning phases in think-aloud protocols, code classroom discourse, and extract patterns from collaborative problem-solving data~\citep{borchers_large_2025,2025.EDM.short-papers.109,zhang_using_2024,vitale_noisy_2025,jiang_uncovering_2025}. Results generally indicate stronger performance when models are given predefined categories or codebooks than when they must generate categories inductively~\citep{DBLP:conf/aied/SimonSTBSBSVACLC25,hou_prompt-based_2024,liu_qualitative_2025,10.1007/978-3-031-76335-9_7}. Together, these studies show that LLMs can apply theory-driven definitions to educational observations, providing a close analogue to interpreting student behavior through learning strategies defined in the literature.

More recent work has extended LLM-based educational analysis beyond text to behavioral and multimodal data. In learning analytics, LLMs have been used to interpret measures derived from interaction data, including quantitative indicators, sequence statistics, and inferred learning processes, and to generate support based on these measures~\citep{jin_chatting_2025,yan_vizchat_2024,wang_self-service_2025,misiejuk2025mapping,opanasenko2025sequence}. For example, \cite{lifloraengine2025} derive self-regulated learning processes and learning conditions from interaction traces using learning analytics and provide them to an LLM to generate adaptive scaffolding. Other work uses fine-tuned multimodal models~\citep{davalos_llms_2025}, structured behavioral representations~\citep{fonteles2026evaluating}, or temporal visualizations and behavioral indicators~\citep{nath2026making,du2026assessment}. These approaches broaden LLM-based interpretation beyond text, but typically transform behavioral traces before presenting them to the model.

Closest to our work, \citet{liu_integrating_2025} examined whether GPT-4o could detect student struggle from gameplay logs converted into human-readable chronological descriptions of student actions, or text replays. GPT-4o was effective when relevant behavioral patterns were not well captured by human-engineered features, but performed less reliably on signals requiring numerical reasoning, such as frequent pauses between actions. The authors therefore combined the LLM with a conventional machine-learning detector, which improved performance on unseen data. However, the study focused on detecting a single behavioral construct within one educational game. In contrast, it remains underexplored whether LLMs can generate evidence-grounded interpretations across multiple theoretically defined learning strategies and learning environments.

\section{Methods}

\begin{figure*}[tp]
    \centering
    \includegraphics[width=0.92\textwidth, page=1]{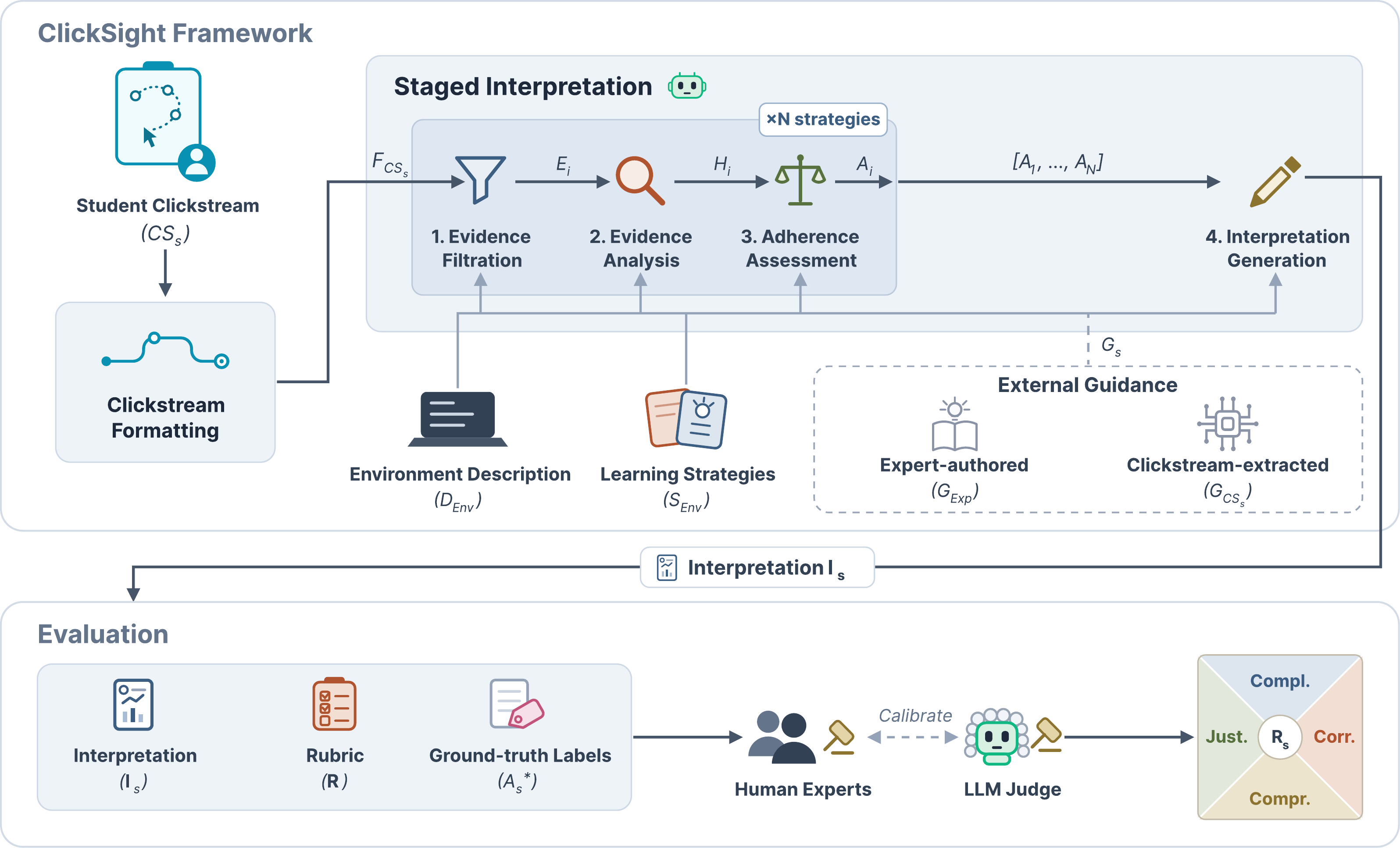}
    \caption{\textbf{Overview of the \cs framework and evaluation process.}
The raw clickstream \(CS_s\) is formatted as \(F_{CS_s}\) and interpreted using the environment description and learning-strategy definitions. For each strategy, three stages filter relevant evidence \(E_i\), analyze it returning \(H_i\), and produce an adherence assessment \(A_i\); the resulting assessments are then combined into the final interpretation \(\mathcal{I}_s\). Optional expert-authored or clickstream-extracted guidance can augment the staged prompts. The interpretation is evaluated with a rubric and ground-truth adherence labels using an LLM judge calibrated with a set of human-expert ratings.}
    \label{fig:pipeline}
\end{figure*}

Our goal is to generate interpretable descriptions of students' learning strategies from clickstream data. To this end, we developed \cs, a framework that transforms raw interaction logs into an LLM-readable representation and uses a staged prompting pipeline to produce evidence-grounded, strategy-based interpretations. We evaluate the resulting interpretations using a rubric-based LLM judge calibrated with a set of human-expert ratings. Figure~\ref{fig:pipeline} provides an overview of the framework and evaluation process.

\subsection{\cs Framework}
\label{sec:framework}

Given a student clickstream, an environment description, and learning-strategy definitions from prior literature, \cs first converts the clickstream into an LLM-readable representation. It then guides the LLM through a four-stage interpretation process developed through consultation with domain experts. The stages are evidence filtration, evidence analysis, adherence assessment, and interpretation generation. The first three stages are applied separately to each strategy, while the final stage combines the resulting strategy-level assessments into a single overall interpretation.

\subsubsection{Inputs and Clickstream Representation}
\label{sec:clickstream_processor}

Let \(CS_s=[e_1,\dots,e_T]\) denote the raw clickstream of student \(s\), where each event \(e_t\) records the student's action, its relevant details, timestamp, and resulting system feedback. Each event is transformed into a standardized symbolic action \(x_t\), producing the formatted sequence \(F_{CS_s}=[x_1,\dots,x_T]\). The transformation retains information relevant to interpretation while removing interface-specific details.

Following prior work by \citet{Radmehr_Singla_Käser_2025}, we represent each formatted action \(x_t\) in a function-like, code-style form:
\begin{center}
\texttt{action\_type(parameters, timestamp)}\\
\texttt{[output: system\_feedback]}
\end{center}

The framework also receives an environment description \(D_{\mathrm{Env}}\) and a predefined set of learning-strategy definitions \(S_{\mathrm{Env}}=\{S_1,\dots,S_N\}\) drawn from prior literature. \(D_{\mathrm{Env}}\) describes the learning goals, available actions, and relevant task structure. Each \(S_i\) defines a learning strategy to be assessed together with its possible adherence levels.
Some environments require separate strategy assessments across multiple dimensions of the task. We therefore define a set of aspects \(K_{\mathrm{Env}}=\{K_1,\dots,K_M\}\), where each aspect specifies a unit over which strategy adherence is evaluated. Each assessment corresponds to a strategy--aspect pair \((S_i,K_j)\). 

\subsubsection{Staged Interpretation Pipeline}
\label{sec:task_descriptions}

\begin{figure*}[tp]
    \centering
    \includegraphics[width=1.0\textwidth, page=2]{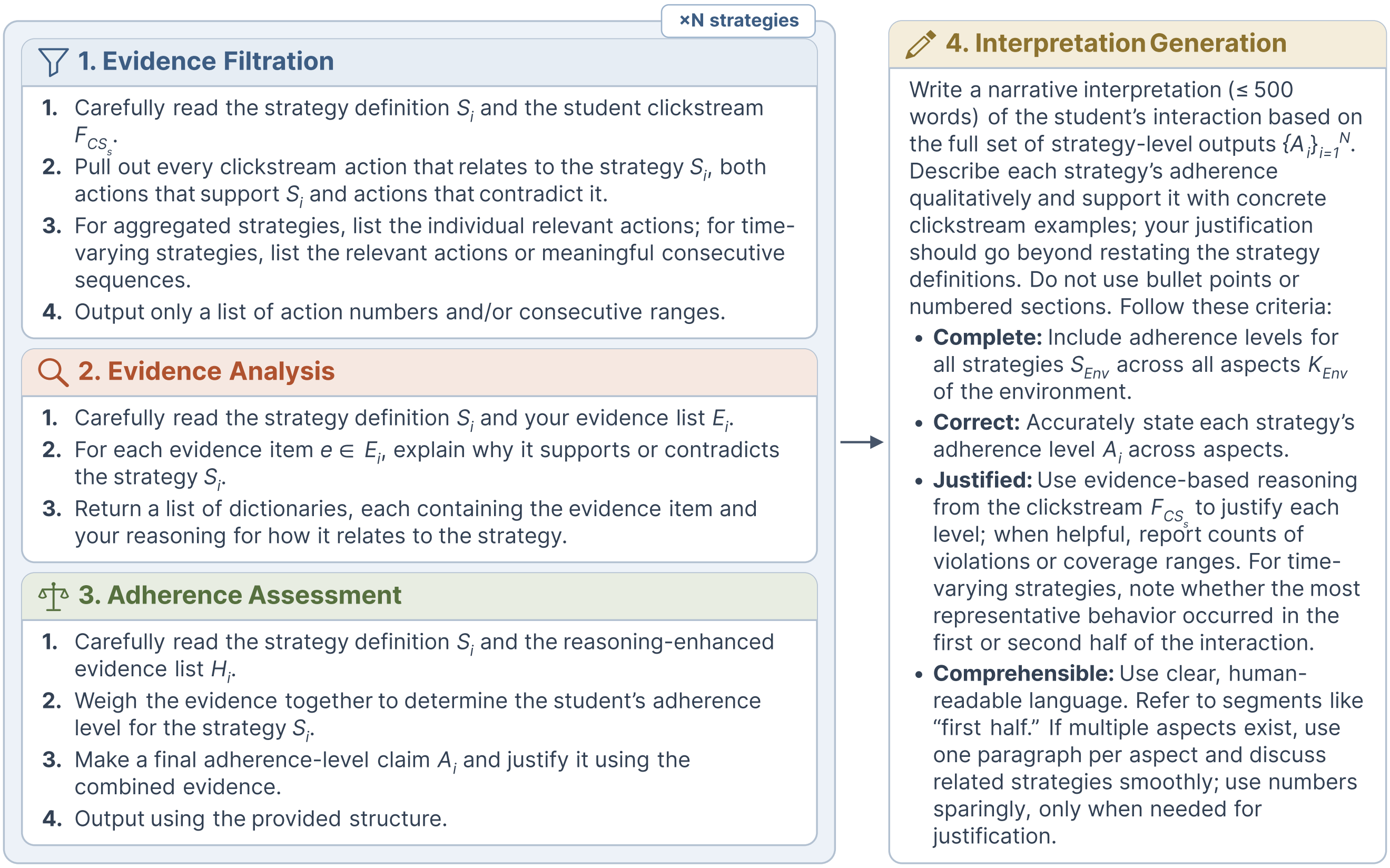}
\caption{\textbf{Staged task descriptions used in the \cs interpretation pipeline.}
Stages~1--3 run independently for each strategy \(S_i\), progressively filtering relevant evidence \(E_i\), analyzing it returning \(H_i\), and producing an adherence assessment \(A_i\). The resulting assessments \(\{A_i\}_{i=1}^{N}\) are then combined in Stage~4 to generate the final narrative interpretation according to the four quality criteria shown.}
    \label{fig:prompts}
\end{figure*}

\cs decomposes clickstream interpretation into four sequential stages. Stages~1--3 run independently for each strategy \(S_i\), while Stage~4 combines the resulting strategy-level assessments. Each stage uses a fixed output schema so that its output can be passed to the next stage. Figure~\ref{fig:prompts} provides the complete task descriptions used at each stage.

For each strategy, \cs progressively narrows the clickstream evidence and converts it into an adherence assessment in Stages~1--3 before synthesizing the strategy-level assessments into the final interpretation in Stage~4:
\begin{enumerate}
    \item \textit{Evidence Filtration.}
    For each strategy \(S_i\), the first stage receives the formatted clickstream \(F_{CS_s}\) and selects actions relevant to assessing adherence across the aspects in \(K_{\mathrm{Env}}\). It produces \(E_i=\{E_{i,j}\}_{K_j\in K_{\mathrm{Env}}}\), where \(E_{i,j}\subseteq F_{CS_s}\) contains the selected evidence for strategy \(S_i\) and aspect \(K_j\).

    \item \textit{Evidence Analysis.}
    The second stage analyzes each selected action with respect to the target strategy. It produces \(H_i=\{H_{i,j}\}_{K_j\in K_{\mathrm{Env}}}\), where each \(H_{i,j}\) pairs the selected evidence with an analysis of what it indicates about adherence to \(S_i\).

    \item \textit{Adherence Assessment.}
    The third stage uses the analyzed evidence to assign an adherence level for each aspect and justify the judgment. It produces \(A_i=\{A_{i,1},\dots,A_{i,M}\}\), where each \(A_{i,j}\) contains an adherence label and an evidence-grounded justification for strategy \(S_i\) and aspect \(K_j\).

    \item \textit{Interpretation Generation.}
    After Stages~1--3 have been completed for all \(N\) strategies, the resulting assessments \(\{A_i\}_{i=1}^{N}\) are provided to the final stage. This stage synthesizes them into a human-readable interpretation \(\mathcal{I}_s\). The prompt also specifies four characteristics of a high-quality interpretation, \compl, \corr, \just, and \compr, which are defined in Section~\ref{sec:eval}.
\end{enumerate}

Only Stage~1 receives the full formatted clickstream. Subsequent stages operate on the output of the preceding stage.

\subsubsection{External Guidance}
\label{sec:guidance}

The core \cs pipeline operates without external guidance. We additionally examine whether augmenting the staged prompts with external guidance improves interpretation quality. We consider two optional forms, expert-authored guidance \(G_{\mathrm{Exp}}\) and clickstream-extracted guidance \(G_{CS_s}\), jointly denoted by \(G_s=[G_{\mathrm{Exp}},G_{CS_s}]\).

Expert-authored guidance \(G_{\mathrm{Exp}}\) consists of either instructions or examples prepared by domain experts familiar with the learning task and strategies being assessed. Instructions provide stage-specific guidance on how a task should be performed, while examples provide sample inputs and outputs for each stage. This guidance is prepared once per environment.

Clickstream-extracted guidance \(G_{CS_s}\) provides structured information derived from the formatted clickstream. We consider indices, which identify where relevant interactions occur, and counts, which summarize how often they occur. Domain experts identify the relevant interaction types once for each environment, such as certain variable explorations. These definitions are then applied \textit{automatically} to each student clickstream to compute the corresponding indices or counts.

\subsection{Evaluation}
\label{sec:eval}

To assess interpretation quality, we developed a rubric grounded in characteristics of high-quality explanations identified in prior work~\citep{rubric1} and refined through consultation with domain experts. The evaluation process is shown at the bottom of Figure~\ref{fig:pipeline}, where generated interpretations are assessed using the rubric together with ground-truth adherence labels derived from the environment-specific detectors used in prior work~(see \Cref{sec:envs}). A set of human-expert ratings is used to calibrate the LLM judge, which is then used to evaluate all generated interpretations.

The rubric consists of four binary criteria, \compl, \corr, \just, and \compr, shown in Table~\ref{tab:ev_cri}. \compl assesses whether the interpretation provides an adherence assessment for a strategy--aspect pair. \corr assesses whether the stated adherence level agrees with the ground-truth label. \just assesses whether the assessment is supported by evidence from the student's clickstream. \compr assesses whether the interpretation communicates the student's behavior clearly in human-readable language. Each criterion is assessed independently for every strategy--aspect pair \((S_i,K_j)\) within an interpretation \(\mathcal{I}_s\). Appendix~\ref{app:ex_d} provides examples of how the four criteria are applied to interpretations.

\begin{table*}[tp]
\centering
\caption{Binary evaluation criteria applied independently to each strategy--aspect pair \((S_i,K_j)\) within a generated interpretation. A score of 0 is assigned when the listed condition is not met.}
\label{tab:ev_cri}
\renewcommand{\arraystretch}{1.3}
\begin{tabularx}{\textwidth}{p{2.5cm} p{5cm} X}
\toprule
\textbf{Criterion} & \textbf{Key Question} & \textbf{Condition for Score = 1} \\
\midrule
\textbf{\compl} &
Does the interpretation provide an adherence assessment for \((S_i,K_j)\)? &
The interpretation contains an assessment of the student's adherence to \((S_i,K_j)\), regardless of correctness.\\
\midrule
\textbf{\corr} &
Does the interpretation correctly assess adherence for \((S_i,K_j)\)? &
The stated adherence level is semantically consistent with the corresponding ground-truth label.\\
\midrule
\textbf{\just} &
Does the interpretation justify the stated adherence assessment for \((S_i,K_j)\) using evidence from the student's clickstream? &
The interpretation provides reasoning that supports the stated adherence assessment, and all referenced actions are present in the clickstream.\\
\midrule
\textbf{\compr} &
Is the interpretation written clearly and grammatically, with natural descriptions of actions and timing? &
The interpretation is grammatically coherent, describes actions in natural language rather than technical action codes, and expresses temporal information using human-readable references such as interaction phases or halves.\\
\bottomrule
\end{tabularx}
\end{table*}

The criteria for \compl, \corr, and \compr are shared across strategies and environments. Assessing \just additionally requires determining whether the cited evidence is appropriate for the specific strategy and adherence level. Domain experts therefore define strategy-specific justification criteria for each environment and strategy. An example of these criteria is provided in \Cref{tab:ex_just_cri}. 

For each interpretation, we compute a score for each criterion as the proportion of strategy--aspect pairs that satisfy that criterion. We additionally report an overall quality score defined as the proportion of strategy--aspect pairs for which all four criteria are satisfied.

We use GPT-5-mini as the LLM judge, which applies the rubric and assigns a binary rating for each criterion and strategy--aspect pair. The prompts for \compl, \corr, and \compr directly ask the judge to assess the corresponding rubric condition. The assessment of \just requires additional verification because evidence can be correctly grounded in the clickstream without actually supporting the stated adherence level. We therefore evaluate \just in three steps. The judge first verifies that the cited evidence is present in the student's interactions, then identifies the adherence assessment stated in the interpretation, and finally determines whether the evidence satisfies the strategy-specific justification requirements for that adherence level. A justification is considered valid only when both the grounding and adherence-specific requirements are satisfied.

We calibrated the LLM judge using a set of human-expert ratings before using it to evaluate the generated interpretations. The calibration procedure is described in Section~\ref{sec:exp_prot}.
\section{Experimental Setup}
\label{sec:exp}

This section describes the two learning environments and datasets used in our experiments, the model backbones and prompting methods used to generate interpretations, and the procedure used to calibrate the LLM judge using human-expert ratings.

\subsection{Environments}
\label{sec:envs}

We evaluated \cs on clickstream data from two educational environments, \ps and \bl (Figure~\ref{fig:envs}), using data from 205 and 160 students, respectively. Both studies received ethics approval from the EPFL Human Research Ethics Committee (HREC; \ps: HREC~062--2021, an amendment of HREC~043--2021; \bl: HREC~2024--13). All participants provided informed consent, and the data were anonymized prior to analysis.

\subsubsection{\textit{\ps}}
\label{sec:ps}

\begin{figure*}[tp]
\centering
\begin{subfigure}[b]{0.48\textwidth}
\centering
{\setlength{\fboxsep}{0pt}\setlength{\fboxrule}{0.3mm}%
\fcolorbox{envframe}{white}{\includegraphics[height=5.34cm]{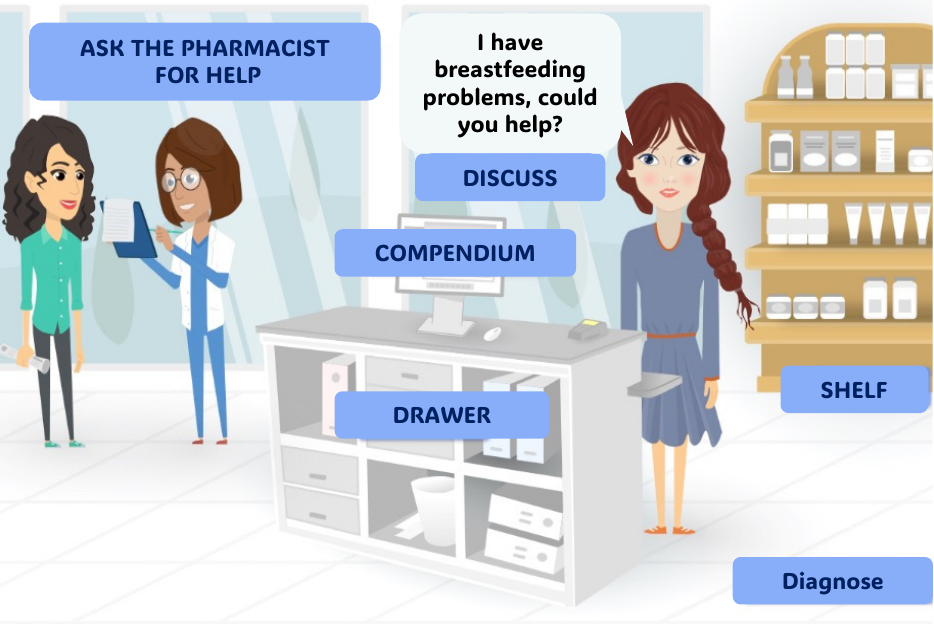}}}
\caption{\ps}
\label{fig:ps}
\end{subfigure}
\hfill
\begin{subfigure}[b]{0.48\textwidth}
\centering
{\setlength{\fboxsep}{0pt}\setlength{\fboxrule}{0.3mm}%
\fcolorbox{envframe}{white}{\includegraphics[height=5.34cm]{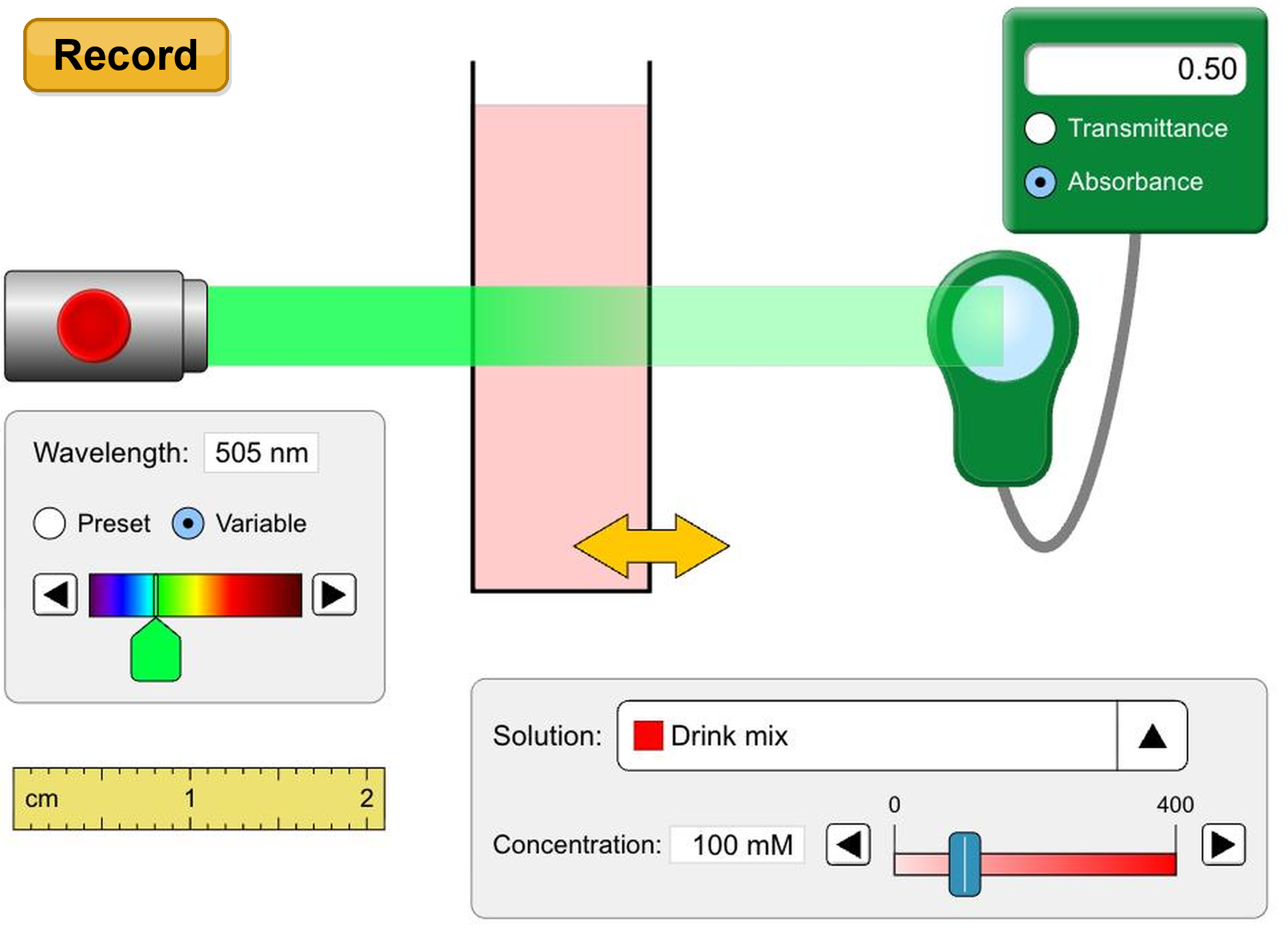}}}
\caption{\bl}
\label{fig:bll}
\end{subfigure}
\caption{\textbf{Educational environments.} \ps (left) is a scenario-based environment for developing diagnostic reasoning through simulated patient interactions. \bl (right) is an interactive PhET simulation for exploring factors affecting light absorbance in solutions.}
\label{fig:envs}
\end{figure*}

\ps (Figure~\ref{fig:ps}) is a scenario-based digital learning environment designed to develop pharmacy assistant trainees' diagnostic reasoning skills~\citep{ps}. Students completed a simulated pharmacy consultation in which they collected information from a virtual patient, consulted reference materials, and selected the most likely diagnosis. The environment automatically logged students' interactions throughout the consultation.

Following prior work~\citep{DBLP:conf/aied/GuresNRRK25,q2,q3}, we consider three diagnostic reasoning strategies:
\begin{enumerate}
    \item \emph{LINDAFF checklist}: whether students ask about a set of clinically relevant topics covering key aspects of the patient's symptoms, such as their intensity, location, and onset. Adherence is binary (\textit{followed} / \textit{not followed}), depending on whether all topics are covered. Table~\ref{tab:ext_info} provides an example of expert-authored guidance for this strategy (see~\Cref{sec:guidance}).

    \item \emph{Interpersonal Relationships}: whether students ask about clinically relevant topics concerning individuals other than the main patient, such as a child, whose condition may matter for diagnosis. Adherence has four levels: \textit{not followed}, \textit{irrelevant inquiry only}, \textit{partial coverage}, and \textit{complete coverage}.

    \item \emph{Broad Diagnostic Reasoning}: whether students gather information relevant to confirming or ruling out all probable diagnoses. Adherence is binary (\textit{followed} / \textit{not followed}), depending on whether all diagnoses are addressed.
\end{enumerate}

We follow the rule-based approach from prior work~\citep{DBLP:conf/aied/GuresNRRK25} to assign ground-truth adherence labels for these strategies, which are used to evaluate the generated interpretations.

\subsubsection{\textit{\bl}}
\label{sec:bll}

\bl (Figure~\ref{fig:bll}) is an inquiry-based PhET simulation in which students manipulate variables such as cuvette width, solution concentration, and laser wavelength to investigate their effects on light absorbance in a chemistry solution~\citep{bll}. The environment automatically logged students' actions together with the relevant simulation state.

Following prior work~\citep{bll}, we consider three inquiry strategies:
\begin{enumerate}
    \item \emph{Control of Variables (CVS)}: whether students isolate the effect of one variable by keeping the others constant. Table~\ref{tab:ex_just_cri} provides an example of expert-defined justification criteria for this strategy (see~\Cref{sec:eval}).

    \item \emph{Exploring a Variable's Full Range}: whether students explore a substantial part of a variable's range and record enough observations to describe its relationship with absorbance.

    \item \emph{Exploring Under Optimal Conditions}: whether students set non-focal variables to values that avoid masking the effect of the focal variable, such as using a nonzero concentration.
\end{enumerate}

Adherence to each strategy is categorized at four levels: \textit{not followed}, \textit{low}, \textit{medium}, or \textit{high}. We use the rule-based approach from prior work~\citep{bll} to assign the corresponding ground-truth adherence labels used in evaluation.

\subsection{Experimental Protocol}
\label{sec:exp_prot}

We compare the \cop pipeline in \cs with two prompting baselines across multiple LLM backbones. We calibrate the LLM judge using a representative set of interpretations independently rated by human experts and assess its agreement with those ratings. Finally, we compare interpretation quality across several conditions using paired statistical tests at the student level.

\subsubsection{Models and Prompting Methods}

We evaluate 12 LLM backbones from three model families: Qwen3, GPT-OSS, and GPT. For each model, we generate interpretations using the \cop pipeline in \cs and two widely used prompting baselines, \zs and \ct. All three methods receive the same formatted clickstream \(F_{CS_s}\), environment description \(D_{\mathrm{Env}}\), and learning-strategy definitions \(S_{\mathrm{Env}}\), differing only in how the model is prompted to construct the interpretation.

In \zs, the model is directly instructed to generate an interpretation satisfying the required quality characteristics, without an explicit decomposition of the reasoning process~\citep{DBLP:journals/corr/abs-2205-11916zeroshot}. In \ct, the model is additionally instructed to reason step by step while constructing the interpretation~\citep{DBLP:conf/nips/Wei0SBIXCLZ22cot}. Its output includes the intermediate reasoning followed by the final interpretation according to a predefined schema. The \cop method instead follows the staged interpretation process described in Section~\ref{sec:task_descriptions}.

The full list of model backbones and decoding settings during inference is provided in Appendix~\ref{app:materials_models}.

\subsubsection{Human Annotation and LLM Judge Calibration}

To calibrate the LLM judge, two learning-science experts independently rated a representative subset of interpretations generated by GPT-5. For \ps, we selected one clickstream from each of six behavioral clusters identified by \(k\)-means clustering of clickstream frequency features, with the number of clusters selected using the elbow method. We generated interpretations for each selected clickstream using \zs, \ct, and \cop, resulting in 18 interpretations. For \bl, we selected one clickstream from each of four behavioral clusters identified in prior work~\citep{bll} and generated interpretations using the same three prompting methods, resulting in 12 interpretations.

The experts independently assessed each interpretation using the rubric in Table~\ref{tab:ev_cri}. Inter-annotator agreement was computed using Cohen's \(\kappa\) separately for each binary criterion. Agreement was perfect for \compl{} and \compr{}, and average \(\kappa\) was at least 0.85 for \corr{} and \just{} in both environments. Full per-criterion values are reported in Table~\ref{tab:app_kappa}.

We use GPT-5-mini as the LLM judge and calibrate it on the same human-rated subset. Each evaluation is done with \(\texttt{temperature}=0.01\) and \(\texttt{top\_p}=0.9\) to obtain a single binary decision for each criterion and strategy--aspect pair. We then assess agreement between the calibrated judge and each human rater using Cohen's \(\kappa\), averaging the resulting values across the two raters. Agreement is high in both environments, with average \(\kappa \geq 0.70\) for every criterion (Table~\ref{tab:app_kappa}). The calibrated judge is then used to evaluate the generated interpretations.

\subsubsection{Statistical Analysis}

For all statistical comparisons, the student is the unit of analysis. We use paired Wilcoxon signed-rank tests for pairwise comparisons and report matched-pairs rank-biserial correlations as effect sizes. Confidence intervals for paired mean differences are obtained by bootstrapping students with 10{,}000 resamples. Holm correction is applied within the predefined families of comparisons for each research question. Full test statistics, confidence intervals, effect sizes, and adjusted $p$-values are reported in Appendix~\ref{app:stats}.%
\section{Results}
\label{sec:results}
We evaluated the quality and consistency of strategy-based interpretations generated by \cs across models and prompting conditions. First, we assessed whether a state-of-the-art model can generate high-quality interpretations without external guidance (RQ1). We then compared interpretation quality across models with different scale and reasoning capabilities (RQ2), and finally, we examined the effect of external guidance (RQ3). Statistical comparisons follow Section~\ref{sec:exp_prot}, with complete results in Appendix~\ref{app:stats}.

\subsection{Quality of state-of-the-art LLM interpretations (RQ1)}
\newcommand{\meanpm}[2]{#1{\tiny\,(\(\pm\)#2)}}

\begin{table*}[tp]
    \centering
    \scriptsize
    \renewcommand{\arraystretch}{1.2}
    \setlength{\tabcolsep}{4pt}
    \caption{\textbf{Quality of GPT-5-generated interpretations.}
    Mean scores with standard errors across prompting methods in \ps and \bl.
    The overall score requires all four criteria to hold for a strategy--aspect pair.
    Superscripts indicate significant improvement over \zs ($^{*}$) or \ct
    ($^{\dagger}$) in Holm-corrected pairwise comparisons
    ($p_{\mathrm{adj}}<.05$).}
    \label{tab:rq1}

    \begin{tabular*}{\linewidth}{@{\extracolsep{\fill}} l l c c c c c}
        \toprule
        \textbf{Environment} & \textbf{Method} &
        \textbf{Completeness} & \textbf{Correctness} &
        \textbf{Justifiedness} & \textbf{Comprehensibility} &
        \textbf{Overall} \\
        \midrule

        \multirow{3}{*}{PharmaSim}
            & \zs
            & 1.000
            & \meanpm{0.958}{0.008}
            & \meanpm{0.977}{0.008}
            & 1.000
            & \cellcolor[HTML]{EEEEEE}\meanpm{0.938}{0.011} \\

            & \ct
            & 1.000
            & \meanpm{0.959}{0.008}
            & \meanpm{0.964}{0.012}
            & 1.000
            & \cellcolor[HTML]{EEEEEE}\meanpm{0.924}{0.014} \\

            & \cop
            & 1.000
            & \meanpm{0.961}{0.008}
            & \meanpm{0.984}{0.009}
            & 1.000
            & \cellcolor[HTML]{EEEEEE}\meanpm{0.945}{0.011} \\

        \midrule

        \multirow{3}{*}{\bl}
            & \zs
            & 1.000
            & \meanpm{0.782}{0.014}
            & \meanpm{0.851}{0.014}$^{\dagger}$
            & 1.000
            & \cellcolor[HTML]{EEEEEE}\meanpm{0.704}{0.016} \\

            & \ct
            & 1.000
            & \meanpm{0.797}{0.012}
            & \meanpm{0.830}{0.014}
            & 1.000
            & \cellcolor[HTML]{EEEEEE}\meanpm{0.702}{0.015} \\

            & \cop
            & 1.000
            & \meanpm{0.815}{0.012}$^{*}$
            & \meanpm{0.844}{0.014}
            & 1.000
            & \cellcolor[HTML]{EEEEEE}\meanpm{0.735}{0.015}$^{*\dagger}$ \\
        \bottomrule
    \end{tabular*}
\end{table*}

We first examined whether GPT-5 can generate high-quality interpretations of student clickstream data. Table~\ref{tab:rq1} reports the scores for \cop and the two baselines, \zs and \ct, across the four evaluation criteria and the overall score. Complete pairwise comparisons are reported in Appendix~\ref{app:rq1}.

Across both environments and all three prompting methods, \compl and \compr reached a perfect score of 1.00. Every interpretation therefore assessed every strategy--aspect pair and expressed the assessment clearly. Differences across conditions were concentrated in \corr and \just.

In \ps, scores were high across all three methods, and no criterion differed significantly by prompting method. In \bl, all methods scored lower on \corr and \just than in \ps. \cop achieved significantly higher \corr than \zs, while neither differed significantly from \ct. For \just, the only significant difference was between the two baselines, with \zs outperforming \ct, while \cop did not differ from either. On the overall score, \cop significantly outperformed both baselines, while \zs and \ct did not differ. Figure~\ref{fig:app_c_methods} in Appendix~\ref{app:ex_c} provides an example of how the three prompting methods interpret the same strategy--aspect pair.

To better understand these differences in \bl, we examined the relationship between \corr and \just. Because \just is assessed against the adherence level stated in the interpretation rather than the ground-truth label, a strategy--aspect pair can satisfy \just while failing \corr. In such cases, the interpretation assigns the wrong adherence level but supports that assessment with evidence from actions the student actually performed. This pattern was particularly visible in \bl, where \just exceeded \corr under all three prompting methods.

The gap between \corr and \just was smallest for \cop, which achieved the highest \corr while maintaining a \just score comparable to both baselines. Since the overall score requires all four criteria to hold for the same strategy--aspect pair, gains in correctness contribute to the overall score only when the interpretation also remains justified. This helps explain why \cop significantly outperformed both baselines overall even though it did not significantly outperform both baselines on \corr or on \just individually.

\vspace{5mm}

\begin{tcolorbox}[
colback=teal!5!white,
colframe=teal!45!black,
boxsep=2pt,
left=6pt,
right=6pt,
top=6pt,
bottom=6pt,
before skip=-2pt,
after skip=2pt]
GPT-5 produced complete and comprehensible interpretations across both environments. Prompting method made little difference in \ps, whereas in \bl{} the \cop{} pipeline achieved the highest overall score and significantly outperformed both baselines.
\end{tcolorbox}

\begin{figure*}[tp]
    \centering
    \includegraphics[width=0.7\textwidth]{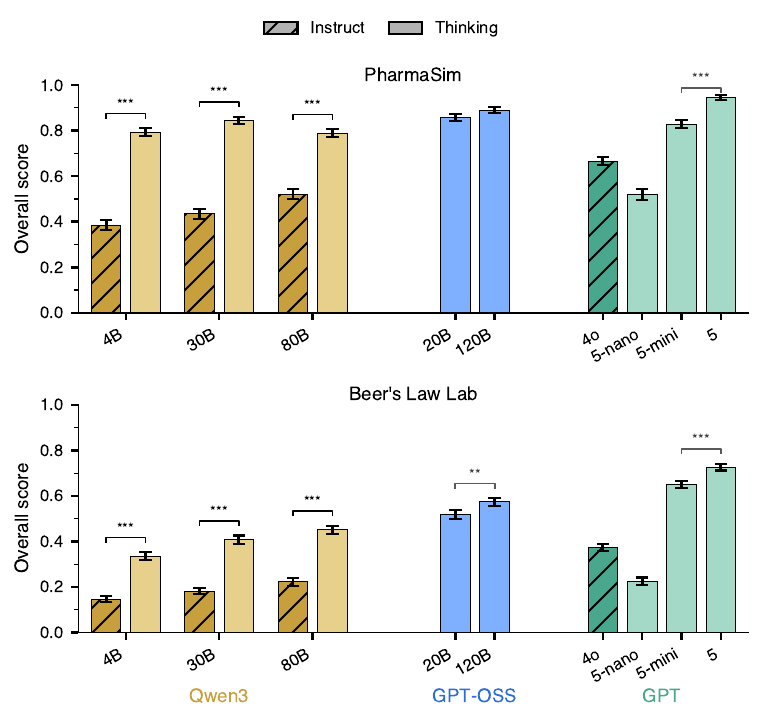}
    \caption{Mean overall scores (including standard errors) across language models on PharmaSim (top) and Beer’s Law Lab (bottom). Results are grouped by model family (Qwen3, GPT-OSS, GPT), with hatched bars indicating instruction-tuned variants and solid bars indicating thinking variants. Brackets mark within-family model contrasts based on Wilcoxon signed-rank tests of per-student scores, Holm-corrected within each family ($^{**}: p<.01$, $^{***}: p<.001$).}
    \label{fig:RQ2}
\end{figure*}

\subsection{Generalization across model capabilities (RQ2)}

We next examined to what extent the findings from GPT-5 generalize across models with different sizes and reasoning capabilities. In particular, we compared the effects of reasoning configuration and model scale. All experiments in this section use \cop, which achieved the highest overall score for GPT-5 in RQ1. Figure~\ref{fig:RQ2} shows the overall scores for the 12 evaluated model configurations in \ps and \bl. Per-criterion scores and complete statistical results are reported in Appendix~\ref{app:rq2}.

Across the Qwen3 family, thinking variants significantly outperformed their instruct counterparts at all three model sizes in both environments. In \ps, switching from instruct to thinking increased the overall score by 0.36 on average across the three model sizes, compared with a 0.21 increase in \bl. For example, Qwen3-4B-Thinking achieved \corr and \just scores of 0.84 and 0.92 in \ps, compared with 0.51 and 0.48 for Qwen3-4B-Instruct (Table~\ref{tab:rq2_criteria_ps}). Appendix~\ref{app:ex_config} provides a matched-size instruct/thinking example for the same student trajectory (Figure~\ref{fig:app_config_pair}).

Model scale showed a less consistent pattern. In \ps, Qwen3-30B-Thinking significantly outperformed the 4B model, with overall scores of 0.85 and 0.79, respectively, while the 80B model also scored 0.79 and did not differ significantly from the 4B model. Because the 80B model is from the Qwen3-Next generation, it differs from the 4B and 30B models in training as well as size, so its lower performance cannot be attributed to model size alone. GPT-OSS similarly showed no significant difference between its 20B and 120B variants.

Among the open-weight thinking models, scaling effects were clearer in \bl. Within the Qwen3 thinking family, performance increased significantly from 4B to 30B. The 80B model scored higher again, although the difference from 30B was not significant. GPT-OSS-120B also significantly outperformed GPT-OSS-20B. In contrast, scaling among the Qwen3 instruct models produced smaller gains than switching to a thinking configuration: increasing model size from 4B to 80B raised the overall score by 0.14 in \ps and 0.08 in \bl.

The GPT family showed a consistent performance ordering across the two environments. GPT-5 achieved the highest overall score, followed by GPT-5-mini. GPT-5-nano, the smallest GPT-5 variant evaluated, performed significantly below GPT-4o in both environments despite using a dedicated reasoning configuration, whereas GPT-4o does not use one. Because these models also differ in size and training, however, this comparison does not isolate the effect of reasoning.

The per-criterion results reveal a further difference between the two environments. For the open-weight thinking models in \ps, \just generally exceeded \corr. In \bl, this pattern reversed. Qwen3-80B-Thinking, for example, achieved 0.81 on \corr but 0.51 on \just, while GPT-OSS-120B achieved 0.77 and 0.69, respectively (Table~\ref{tab:rq2_criteria_bll}). The gap was particularly large for Qwen3, where \corr exceeded \just by approximately 0.3 across the three thinking variants. GPT-5 did not show the same pattern, achieving similarly high scores on the two criteria.

Thus, the open-weight models in \bl were more successful at assigning the correct adherence level than at providing the evidence required to justify that assessment. Because \just in \bl requires exact exploration counts and information about where those explorations occur in the interaction, this pattern motivates RQ3, where we test whether supplying such information directly improves interpretation quality.

\vspace{5mm}

\begin{tcolorbox}[
colback=teal!5!white,
colframe=teal!45!black,
boxsep=2pt,
left=6pt,
right=6pt,
top=6pt,
bottom=6pt,
before skip=-2pt,
after skip=2pt]
Reasoning configuration was more consistently associated with interpretation quality than model scale. Scaling was less consistent in \ps but provided clearer gains in \bl.
\end{tcolorbox}

\begin{figure*}[tp]
\centering
\includegraphics[width=0.7\linewidth]{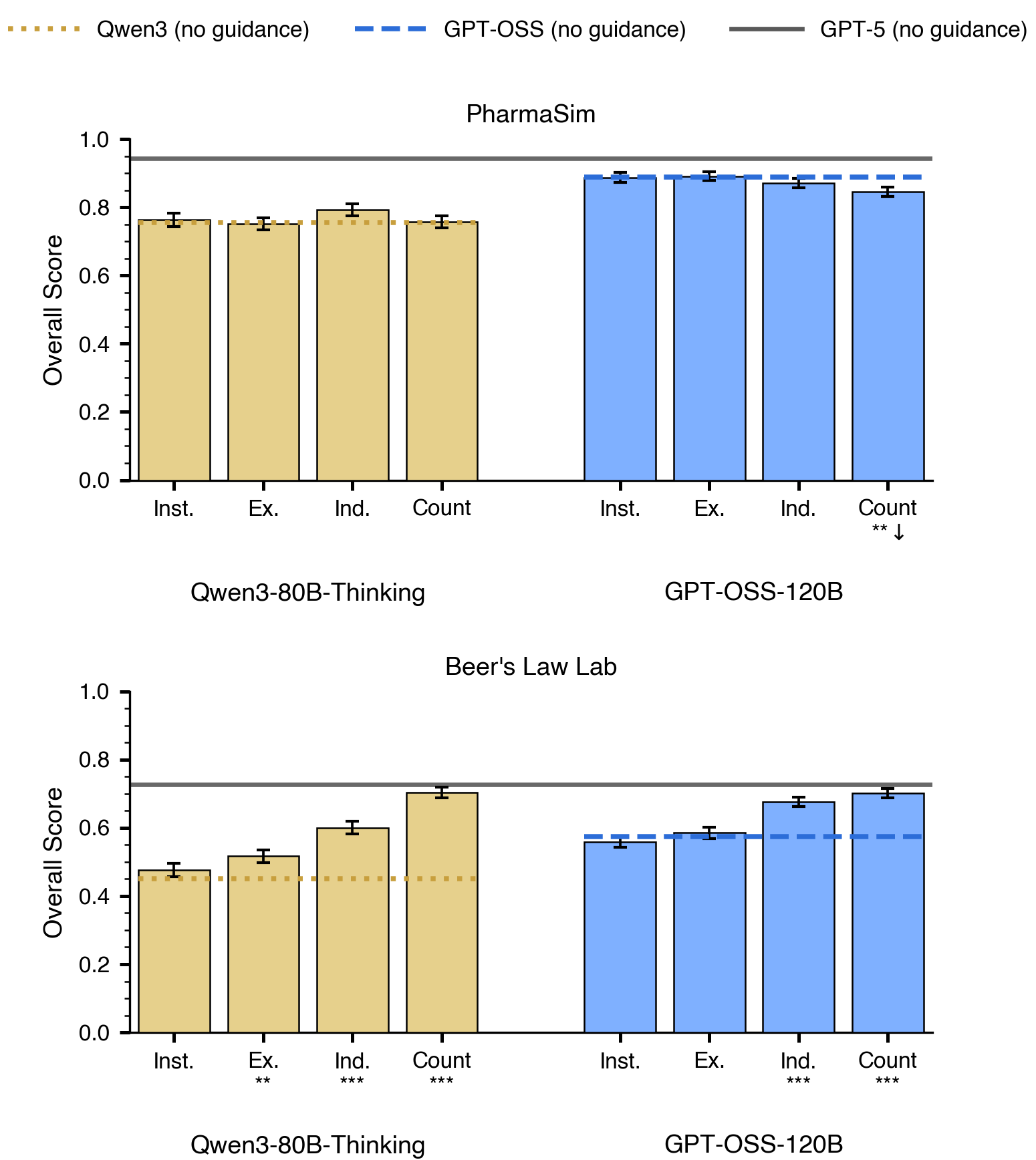}
\caption{\textbf{Effect of auxiliary guidance on overall performance.} Bars show Qwen3-80B-Thinking and GPT-OSS-120B on \ps (top) and \bl (bottom) with additional instructions (Inst.), five examples (Ex.), interaction indices (Ind.), or interaction counts (Count). Dotted and dashed lines mark each model's no-guidance baseline; the solid grey line marks GPT-5 without auxiliary guidance. Error bars are standard errors over students. Marks beneath each condition indicate significant differences from the same model's no-guidance baseline based on Holm-corrected pairwise comparisons ($^{*}p_{\mathrm{adj}}<.05$, $^{**}p_{\mathrm{adj}}<.01$, $^{***}p_{\mathrm{adj}}<.001$). A downward arrow denotes a significant decrease.}

\label{fig:ext_info}
\end{figure*}

\subsection{Impact of External Guidance (RQ3)}

Finally, we examined whether external guidance improves interpretation quality. We compared expert-authored instructions (Inst.) and examples (Ex.) with clickstream-extracted indices (Ind.) and counts (Count) using Qwen3-80B-Thinking and GPT-OSS-120B with \cop. Results are shown in Figure~\ref{fig:ext_info}, with per-criterion scores and complete statistical comparisons reported in Appendix~\ref{app:rq3}.

In \ps, none of the guidance conditions significantly improved either model. Interaction indices produced the largest positive change for Qwen3-80B-Thinking, but the difference was not significant. The only significant change was negative, with counts reducing GPT-OSS-120B's overall score by 0.04.

In \bl, both forms of clickstream-extracted guidance significantly improved both models, with counts producing the largest gains. Qwen3-80B-Thinking increased from 0.45 without guidance to 0.60 with action indices and 0.70 with action counts, while GPT-OSS-120B increased from 0.57 to 0.68 and 0.70, respectively. These gains came primarily from \just rather than \corr. With counts, Qwen3-80B-Thinking's \just increased from 0.51 to 0.83 while \corr remained at 0.81--0.82, and GPT-OSS-120B's \just increased from 0.69 to 0.87 while \corr remained around 0.77. Indices showed the same pattern, but with smaller gains. With counts, both models also came within approximately 0.03 of GPT-5, with no significant remaining difference.

Expert-authored guidance produced smaller effects. Additional instructions did not significantly improve either model in either environment. Example-based guidance significantly improved only Qwen3-80B-Thinking in \bl, increasing its overall score by 0.07, compared with a 0.25 increase from action-count guidance. GPT-OSS-120B did not significantly benefit from example-based guidance. In aggregate, clickstream-extracted guidance significantly outperformed expert-authored guidance for both models in \bl.

Appendix~\ref{app:ex_guidance} illustrates the effect of action-count guidance with a before-and-after example (Figure~\ref{fig:app_guidance_pair}). Appendix~\ref{app:length} further examines how interpretation quality varies with clickstream length. In \bl, longer traces were associated with lower overall scores, and this decline was weaker under action-count guidance for both models. The corresponding patterns in \ps were smaller and inconsistent.

\vspace{2mm}
\begin{tcolorbox}[
colback=teal!5!white,
colframe=teal!45!black,
boxsep=2pt,
left=6pt,
right=6pt,
top=6pt,
bottom=6pt,
before skip=-2pt,
after skip=2pt]
External guidance had different effects across the two environments. In \bl, clickstream-extracted guidance improved both open-weight models, bringing their overall scores within 0.03 of GPT-5, while expert-authored guidance produced smaller gains. In \ps, none of the guidance conditions improved performance.
\end{tcolorbox}

\section{Discussion}
\label{sec:discussion}
This study examined whether LLMs can connect low-level student clickstream data with theoretically defined learning strategies. Across two distinct learning environments, our findings show that LLMs can generate meaningful strategy-based interpretations, but that their success depends on the reasoning demands of the strategy, the structure of the prompting approach, and the information that can be reliably recovered from the behavioral trace. These findings reveal both the potential and the limitations of LLMs for interpreting behavioral learning data and point toward a complementary role for LLM-based interpretation and established learning analytics methods.

\subsection{Strategy demands and interpretation quality}

The two environments differed substantially in both interpretation quality and the effect of prompting. In \ps, all three prompting methods achieved similarly high scores, whereas in \bl, \corr and \just were lower and \cop significantly outperformed both baselines overall. One likely reason is the type of evidence that must be recovered from the trace. The diagnostic strategies in \ps are largely time-invariant and often depend on whether particular actions occurred, such as asking about a relevant symptom~\citep{DBLP:conf/aied/GuresNRRK25,ps}. Control of Variables and related strategies in \bl instead require the model to compare actions across the interaction, determine what changed and what remained constant, and reconstruct how the exploration unfolded over time~\citep{perez2018control,bll}.

The number of adherence levels alone does not explain this difference. Interpersonal Relationships in \ps has four adherence levels, as do the strategies in \bl, but performs similarly to the other \ps strategies. Within \bl, Control of Variables has the lowest scores and also requires exact exploration counts and comparisons across actions. What seems to matter more is how much sequential and quantitative evidence must be reconstructed before an adherence level can be assigned. The full per-strategy analysis is reported in Appendix~\ref{app:per_strategy}.

This also helps explain why the prompting methods behaved differently across the two environments. In \bl, asking the model to reason step by step with \ct did not improve the overall score over \zs and resulted in significantly lower \just. \cop instead separates evidence selection, analysis, and adherence assessment before the final interpretation is generated. This kind of decomposition has been shown to help on multi-step tasks~\citep{sun-etal-2024-promptchain,DBLP:conf/nips/Wei0SBIXCLZ22cot}, and here its advantage appears when the model must first recover relevant evidence from a sequence of actions and then interpret that evidence through a learning strategy. In \ps, where much of the required evidence can be identified more directly, the differences between prompting methods were much smaller, consistent with prior findings that step-by-step prompting offers limited benefit when a task can already be solved directly~\citep{DBLP:journals/corr/abs-2205-11916zeroshot}.

A separate problem concerns the relationship between \corr and \just. An interpretation could assign the wrong adherence level while still supporting that judgment with actions that genuinely appeared in the clickstream. In these cases, the issue was not fabricated evidence. The model had selected real evidence, but not necessarily all of the evidence needed to reach the correct judgment.

This makes such errors difficult to detect from the interpretation alone. Verifying that the cited actions occurred establishes that the interpretation is grounded in the trace, but not that the cited evidence is sufficient. Relevant actions may still have been omitted. This connects to a broader concern that clear explanations can support incorrect conclusions~\citep{rubric1,rubric2}. A similar problem has been reported for LLM interpretations of temporal learner-data visualizations, where coherent and theoretically plausible accounts are not always supported by the underlying data~\citep{nath2026making}. For clickstream interpretation, justification therefore depends not only on whether the cited actions are real, but also on whether the interpretation accounts for all the evidence needed to support or challenge the stated adherence level.

\subsection{Reasoning capability and its limits}

Reasoning configuration was more consistently associated with interpretation quality than model scale. Within the Qwen3 family, thinking variants outperformed their matched instruct counterparts at every evaluated size in both environments, and the gains from switching to a thinking configuration were larger than those obtained by scaling the instruct models. This is consistent with recent educational applications in which reasoning-oriented models performed better on tasks requiring multi-step interpretation~\citep{schiott2025using,du2026assessment}. Model scale showed a less consistent pattern. Scaling improved performance in some \bl conditions but not consistently in \ps. One possible explanation is that the longer clickstreams in \bl place greater demands on a model's ability to retain and integrate information across the interaction.

The criterion-level results also show that stronger reasoning does not address all sources of error. In \ps, \just generally exceeded \corr for the open-weight thinking models. In \bl, the pattern reversed: models often assigned the correct adherence level but failed to provide the precise evidence required to justify it, with the largest gaps observed for Qwen3. This suggests that reasoning over identified evidence and accurately recovering that evidence from the clickstream are distinct challenges.

Errors may therefore arise at different stages of the interpretation process. A model may recover the relevant evidence but interpret it incorrectly, or it may fail to recover the required evidence from the clickstream in the first place. Stronger reasoning may help address the former without necessarily resolving the latter. This distinction motivates RQ3, where we examine whether directly providing relevant quantitative information from the clickstream improves interpretation quality.

\subsection{Grounding LLM interpretations with quantitative trace information}

The external-guidance results show that deterministic trace analysis can complement LLM-based interpretation. In \bl, both action indices and action counts improved the two open-weight models, with action counts producing the largest gains. The improvements were concentrated in \just, while \corr remained nearly unchanged. The additional information therefore did not substantially alter the models' adherence judgments, but improved their ability to support those judgments with the quantitative evidence required by the evaluation criteria.

Expert-authored guidance had a smaller effect. Additional instructions did not significantly improve either model, and example-based guidance improved only Qwen3-80B-Thinking in \bl. This differs from LLM-based coding of educational text, where performance often depends strongly on the quality of definitions and examples~\citep{borchers_large_2025,zhang_using_2024,hou_prompt-based_2024,10.1007/978-3-031-76335-9_7,DBLP:conf/aied/SimonSTBSBSVACLC25,liu_qualitative_2025}. In our setting, the strategy definitions were already explicit. The main difficulty was therefore often not interpreting the strategy definition, but recovering the precise information required to apply it to a long behavioral sequence.

The exploratory trace-length analysis provides additional evidence for this explanation. In \bl, longer clickstreams were associated with lower overall scores, and this decline was weaker when action-count guidance was provided. In \ps, the corresponding patterns were smaller and inconsistent. Although this analysis does not establish a causal effect of trace length, it suggests that recovering relevant information may become more difficult as the amount of behavioral information increases. The full trace-length analysis is reported in Appendix~\ref{app:length}.

Together, these results motivate a hybrid approach that combines LLM-based interpretation with established learning-analytics methods. Educational data mining provides reliable methods for extracting quantities, sequences, transitions, and temporal patterns from behavioral 
traces~\citep{FRA17,DBLP:conf/lak/ParkDRSW17,DBLP:conf/lak/SaqrLTKMT25,DBLP:conf/lak/ChengGLR0FJTGS25,lifloraengine2025}. Such methods can recover information about order, frequency, and timing, while the LLM can relate this information to theoretically defined strategies and produce a human-readable interpretation. The two components therefore address different parts of the interpretation process.

This hybrid approach also changes where expert input is required. Experts still define the strategies and identify which behavioral quantities are relevant, but they need not encode the complete mapping from behavioral traces to strategy interpretations or provide labeled examples for individual students. Our experiments considered only simple action indices and counts. Future work could examine richer trace-derived information, such as sequential patterns or transition measures.

\subsection{Implications for research and practice}
\label{sec:implications}

For research, \cs provides a way to examine theory-defined learning strategies at the level of individual sessions without requiring labeled data or a dedicated detector for each strategy. Researchers can provide the strategy definition and examine how the model relates it to the behavioral trace, rather than first encoding the strategy as a task-specific set of rules or training examples. This may make it easier to study the same theoretical construct across different learning environments while reducing the need to specify in advance how that construct should appear in each setting.

The evidence associated with each judgment can also support closer inspection of the interpretation. Rather than returning only an adherence label, \cs identifies the actions used to support that label. Researchers can compare these actions with theoretical expectations and with existing quantitative measures. However, the presence of cited actions in the trace does not by itself establish that the evidence is sufficient, since relevant actions may still have been omitted.

This is particularly important to be considered for teacher-facing applications. Interpretations should be presented as judgments that can be examined rather than as established facts about a student. Interfaces could expose the supporting actions, help users inspect potentially omitted evidence, and flag cases in which models or methods disagree. This is consistent with work arguing that learning-analytics tools should support rather than replace professional judgment~\citep{DBLP:conf/lak/Wang0LN24,molenaar2022concept}. The present study, however, evaluates interpretation quality rather than the effect of these interpretations on teacher decision-making.

Finally, strategy adherence should also not be treated as a stable characteristic of a student. The judgment concerns behavior observed in a particular session and environment, and similar observable behavior may arise for different reasons~\citep{baker2006gaming,DBLP:conf/lak/KimLJX25}. The interpretation therefore characterizes the observed learning process rather than the learner.

\subsection{Limitations and Future Work}
\label{sec:limitations}

Our results come from two learning environments, one cohort per environment, and six learning strategies. Both settings involve a single student completing a relatively short activity, so the findings may not extend to collaborative or long-running learning environments. We further compared \cs only with prompting baselines and did not evaluate fine-tuned or supervised alternatives. Future work should broaden the range of environments and strategies and compare \cs systematically with alternative approaches while accounting for both setup and inference costs.

A second set of limitations concerns the scope of the evaluation and its implications for instructional use. This study focuses on interpretation quality as a necessary first step toward instructional use. Future work should examine how teachers understand and use these interpretations, how these interpretations compare with existing representations such as dashboards, and whether they support instructional decisions. Our evaluation also reports aggregate performance, leaving open whether interpretation quality varies across demographic or other groups of students and whether systematic biases exist in the resulting interpretations.

The current implementation of \cs transforms raw log events into a function-like, LLM-readable representation. This representation makes the behavioral trace explicit and consistent for the model, but other possible transformations remain unexplored. Future work should examine how different representations affect interpretation quality, how much of the transformation can be automated, and whether models can work effectively with less structured or raw event logs. Similarly, our experiments with external guidance considered only simple action indices and counts. Extending this guidance to richer quantitative information, such as sequential patterns or transition-based measures, could help identify which forms of trace processing are most useful to LLMs for interpreting different types of learning strategies.

Together, these directions can help establish when LLMs can reliably connect behavioral traces to theoretically meaningful learning strategies and when they require complementary computational or human support.

\section*{Declaration of generative AI and AI-assisted technologies in the manuscript preparation process}

During the preparation of this work, the authors used OpenAI's ChatGPT and Anthropic's Claude to shorten and polish the writing, and to generate the vector icons used in Figures~\ref{fig:pipeline} and~\ref{fig:prompts}.
After using these tools, the authors reviewed and edited the content as needed and take full responsibility for the content of the published article.
\section*{Code and data availability}
The \cs implementation, prompts, judge rubrics, the sampling procedures and seeds
behind every example in Appendix~\ref{app:ex_catalogue}, and the scripts that
generate all appendix tables and figures are available at \url{https://github.com/epfl-ml4ed/ClickSight-CEAI}.
Student identifiers are anonymized throughout; the raw interaction data
cannot be shared publicly under the terms of the underlying studies.

\section*{Acknowledgements}
This project was substantially funded by the Swiss State Secretariat for
Education, Research and Innovation (SERI). We thank Fares Fawzi for his valuable feedback.

\section*{Declaration of competing interests}
The authors have no relevant financial or non-financial interests to disclose.

\section*{CRediT authorship contribution statement}
\textbf{Bahar Radmehr:} Conceptualization, Methodology, Software, Formal
analysis, Data curation, Investigation, Visualization, Writing -- original draft.
\textbf{Lan Tran:} Software, Investigation.
\textbf{Ekaterina Shved:} Methodology, Data curation, Validation.
\textbf{Fatma Bet\"ul G\"ure\c{s}:} Methodology, Validation.
\textbf{Adish Singla:} Conceptualization, Methodology, Supervision, Writing -- review and editing.
\textbf{Tanja K\"aser:} Conceptualization, Methodology, Supervision, Funding
acquisition, Project administration, Writing -- review and editing.
\label{LastMainPage}
\newpage
\pagestyle{empty}

\bibliographystyle{cas-model2-names}
\bibliography{bibliography}

\begin{appendices}

\raggedbottom
\renewcommand{\arraystretch}{1.15}

\counterwithin{table}{section}
\counterwithin{figure}{section}
\counterwithin{equation}{section}

\section{Materials and implementation details}
\label{app:materials}

This appendix provides the materials and implementation details referenced in the Methods and Experimental Setup. It includes examples of the expert-authored guidance used in RQ3, an example of the strategy-specific criteria used to evaluate \just{}, the evaluated model backbones and decoding settings, and the full inter-rater agreement results.

\subsection{Expert-authored guidance and justification criteria}
\label{app:materials_guidance}

Table~\ref{tab:ext_info} shows the expert-authored instructions and examples provided for the LINDAFF checklist in \ps{} at each stage of the \cop{} pipeline. Table~\ref{tab:ex_just_cri} provides the justification criteria for one strategy, Control of Variables in \bl{}, to illustrate how \just{} is evaluated for different adherence levels.

\begin{table*}[tp]
\centering
\caption{Example expert-authored guidance for the LINDAFF checklist at each stage of the \cop{} pipeline. \textit{Instruction} specifies how to complete the stage, and \textit{Examples} show how the instruction is applied to a student interaction.}
\label{tab:ext_info}

\begin{tabularx}{\textwidth}{>{\RaggedRight\arraybackslash}m{2.2cm} >{\RaggedRight\arraybackslash}X >{\RaggedRight\arraybackslash}X}
\toprule
\textbf{Stage} & \textbf{Instruction} & \textbf{Examples} \\
\hline
\rowcolor{lightpink}
\textbf{Evidence Filtration} &
Filter the actions that ask questions about the patient. &
\texttt{[discuss(mother, symptoms), discuss(baby, symptoms), ...]}
\medskip
$\rightarrow$\texttt{[discuss(mother, symptoms), discuss(mother, diet)]} \\
\hline
\rowcolor{lightgreen}
\textbf{Evidence Analysis} &
For each piece of evidence (i.e., each question asked about the patient), check if it corresponds to one of the checklist topics. &
\texttt{[discuss(mother, symptoms), discuss(mother, diet)]}
\medskip
$\rightarrow$\texttt{[discuss(mother, symptoms) covers nature of symptoms,
discuss(mother, diet) covers worsening factors]} \\
\hline
\rowcolor{lightblue}
\textbf{Adherence Assessment} &
If the evidence shows that all LINDAFF topics were addressed, mark the adherence level as \textit{followed}. Otherwise, mark it as \textit{not followed}.
\medskip
Justify the adherence level for:
- \textit{Not Followed}: Clearly state which LINDAFF topics were missing, or list only the subset covered.
- \textit{Followed}: Clearly state that all components were addressed and support with examples. &
\texttt{[discuss(mother, symptoms) covers nature of symptoms,
discuss(mother, diet) covers worsening factors]}
\medskip
$\rightarrow$ Not followed: The student did not cover intensity, location, duration, medical history, or relieving factors. \\
\hline
\rowcolor{lightsalmon}
\textbf{Interpretation Generation} &
Ensure the interpretation meets the required criteria and is supported with evidence\dots &
\texttt{[discuss(mother, symptoms), discuss(baby, symptoms), ...]}
\medskip
$\rightarrow$ The student did not follow the LINDAFF checklist because they did not address symptom intensity, location, duration, relevant medical history, or relieving factors. \\
\hline
\end{tabularx}
\end{table*}

\begin{table*}[tp]
\centering
\caption{Example justification criteria for the Control of Variables strategy. A \just{} score of 1 is assigned only when the listed conditions for the stated adherence level are satisfied.}
\label{tab:ex_just_cri}
\renewcommand{\arraystretch}{1.25}

\begin{tabularx}{\textwidth}{>{\RaggedRight\arraybackslash}p{3.8cm} X}
\toprule
\textbf{Adherence Level} &
\textbf{Condition for Score = 1 (Justified)} \\
\midrule
\textbf{Not Followed} &
The interpretation explicitly states that the student \textbf{did not explore} the variable associated with the aspect at all. \\
\midrule
\textbf{Low / Medium / High} &
The interpretation specifies all of the following:
\vspace{-0.4em}
\begin{enumerate}[leftmargin=1.4em, itemsep=0.15em, topsep=0.2em]
    \item \textbf{Number of explorations}: how many explorations of the variable related to the aspect were performed.
    \item \textbf{Adherence breakdown}: how many, or what percentage, of those explorations followed the strategy.
    \item \textbf{Half of the interaction}: whether most explorations occurred in the first or second half of the interaction.
\end{enumerate} \\
\bottomrule
\end{tabularx}
\end{table*}

\subsection{Model backbones and decoding settings}
\label{app:materials_models}

We evaluated twelve model configurations from three families. The Qwen3 configurations comprise 4B, 30B, and 80B models in both instruct and thinking variants~\citep{yang2025qwen3technicalreport}. We also evaluated GPT-OSS-20B and GPT-OSS-120B~\citep{openai2025gptoss}, and the proprietary OpenAI models GPT-4o, GPT-5-nano, GPT-5-mini, and GPT-5~\citep{openai2025gpt5}. For generation, we used \(\texttt{temperature}=0.1\) and \(\texttt{top\_p}=0.9\). Open-weight models were run with vLLM~\citep{kwon2023efficient}.

\subsection{Inter-rater agreement}
\label{app:materials_kappa}

Table~\ref{tab:app_kappa} reports the agreement statistics summarized in Section~\ref{sec:exp_prot}. Human--human agreement is reported between the two expert raters, and LLM--human agreement is computed between the calibrated LLM judge and those ratings.

\begin{table}[tp]
\centering
\caption{Inter-rater agreement (Cohen's $\kappa$) for the rubric criteria. Human--human values are averaged across strategy--aspect-pair \(\kappa\) values, with the minimum shown in parentheses. LLM--human values are averaged across the two human raters.}
\label{tab:app_kappa}
\footnotesize
\setlength{\tabcolsep}{3pt}
\begin{tabular}{@{}l l c c@{}}
\toprule
\textbf{Environment} & \textbf{Criterion} &
  \makecell{\textbf{Human--human}\\ $\kappa_{\text{avg}}$ ($\kappa_{\min}$)} &
  \makecell{\textbf{LLM--}\\\textbf{human} $\kappa$} \\
\midrule
\multirow{4}{*}{\ps{}}
  & \compl & 1.00 (1.00) & 1.00 \\
  & \corr  & 1.00 (1.00) & 1.00 \\
  & \just  & 0.85 (0.75) & 0.70 \\
  & \compr & 1.00 (1.00) & 1.00 \\
\midrule
\multirow{4}{*}{\bl}
  & \compl & 1.00 (1.00) & 1.00 \\
  & \corr  & 0.98 (0.82) & 0.95 \\
  & \just  & 0.90 (0.82) & 0.95 \\
  & \compr & 1.00 (1.00) & 1.00 \\
\bottomrule
\end{tabular}
\end{table}

\FloatBarrier

\section{Detailed statistical results}
\label{app:stats}

This appendix reports the per-criterion scores and complete statistical comparisons supporting the results in the main text. The student is the unit of analysis. Because each student contributes multiple strategy--aspect pairs, pairwise tests are conducted over student-level scores. Confidence intervals for paired mean differences use a cluster bootstrap with 10{,}000 student-level resamples. Score tables report means across students with standard errors in parentheses.

Comparisons are conducted separately within \ps{} and \bl{} because the environments contain different strategies and label spaces. Paired comparisons use Wilcoxon signed-rank tests with matched-pairs rank-biserial correlations as effect sizes. Holm correction is applied within the comparison families identified in each table. Two correction families span both environments: thinking versus instruct across the three Qwen3 sizes, and clickstream-extracted versus expert-authored guidance across two models and three outcomes. The overall score is the conjunction of all four criteria for a strategy--aspect pair, rather than their mean, and can therefore be lower than each individual criterion. In the score tables, values shown without a standard error have zero variation across students. Where bold is used, it marks a column maximum that significantly exceeds the second-highest value under the corresponding paired comparison and Holm correction.

\subsection{RQ1: prompting strategies}
\label{app:rq1}

RQ1 compares \cop{}, \zs{}, and \ct{} with GPT-5 while holding the model and input data fixed. Table~\ref{tab:rq1_contrasts} reports the pairwise comparisons for \corr{}, \just{}, and the overall score. \compl{} and \compr{} are omitted because both equal 1.00 for all three prompting methods in both environments, leaving no variation to test. Holm correction is applied to the three method comparisons within each criterion and environment ($k=3$).

No prompting-method comparison is significant in \ps{}. In \bl{}, \cop{} scores significantly higher than \zs{} on \corr{} and significantly higher than both \zs{} and \ct{} on the overall score. For \just{}, \zs{} scores significantly higher than \ct{}, while \cop{} does not differ significantly from either baseline. These are the comparisons marked by superscripts in Table~\ref{tab:rq1}.

\begin{table*}[tp]
\centering
\begin{threeparttable}
\caption{RQ1 pairwise contrasts between prompting strategies.
         \ps{} $n = 205$ students (upper block), \bl{} $n = 160$ (lower block). $\Delta$ is the paired mean difference and $r_{\text{rb}}$ the
         matched-pairs rank-biserial correlation. Bold $p_{\text{Holm}}$ marks $p < .05$ after
         Holm correction within each family of three contrasts.}
\label{tab:rq1_contrasts}
\small
\setlength{\tabcolsep}{3pt}
\begin{tabular*}{\linewidth}{@{\extracolsep{\fill}} l l r l r c c}
\toprule
\textbf{Criterion} & \textbf{Contrast} & $\Delta$ & \textbf{95\% CI} &
  $r_{\text{rb}}$ & $p$ & $p_{\text{Holm}}$ \\
\midrule
\multicolumn{7}{l}{\textit{\ps}} \\
\midrule
\multirow{3}{*}{\corr} & \cop $-$ \zs & $+0.003$ & $[-0.015, +0.023]$ & $+0.06$ & .732 & $>$.999 \\
 & \cop $-$ \ct & $+0.002$ & $[-0.016, +0.020]$ & $+0.03$ & .853 & $>$.999 \\
 & \ct $-$ \zs & $+0.002$ & $[-0.015, +0.018]$ & $+0.04$ & .841 & $>$.999 \\
\midrule
\multirow{3}{*}{\just} & \cop $-$ \zs & $+0.007$ & $[-0.016, +0.031]$ & $+0.19$ & .534 & .785 \\
 & \cop $-$ \ct & $+0.020$ & $[-0.010, +0.050]$ & $+0.36$ & .262 & .785 \\
 & \ct $-$ \zs & $-0.013$ & $[-0.042, +0.015]$ & $-0.25$ & .361 & .785 \\
\midrule
\multirow{3}{*}{Overall} & \cop $-$ \zs & $+0.007$ & $[-0.023, +0.036]$ & $+0.10$ & .546 & .831 \\
 & \cop $-$ \ct & $+0.021$ & $[-0.011, +0.055]$ & $+0.19$ & .253 & .758 \\
 & \ct $-$ \zs & $-0.015$ & $[-0.047, +0.016]$ & $-0.14$ & .416 & .831 \\
\midrule
\multicolumn{7}{l}{\textit{\bl}} \\
\midrule
\multirow{3}{*}{\corr} & \cop $-$ \zs & $+0.033$ & $[+0.009, +0.056]$ & $+0.33$ & .006 & \textbf{.018} \\
 & \cop $-$ \ct & $+0.017$ & $[-0.003, +0.038]$ & $+0.22$ & .082 & .164 \\
 & \ct $-$ \zs & $+0.015$ & $[-0.005, +0.035]$ & $+0.18$ & .151 & .164 \\
\midrule
\multirow{3}{*}{\just} & \cop $-$ \zs & $-0.008$ & $[-0.019, +0.004]$ & $-0.27$ & .197 & .200 \\
 & \cop $-$ \ct & $+0.014$ & $[-0.002, +0.031]$ & $+0.26$ & .100 & .200 \\
 & \ct $-$ \zs & $-0.022$ & $[-0.039, -0.006]$ & $-0.43$ & .011 & \textbf{.032} \\
\midrule
\multirow{3}{*}{Overall} & \cop $-$ \zs & $+0.031$ & $[+0.011, +0.050]$ & $+0.36$ & .004 & \textbf{.007} \\
 & \cop $-$ \ct & $+0.033$ & $[+0.011, +0.055]$ & $+0.39$ & .002 & \textbf{.006} \\
 & \ct $-$ \zs & $-0.002$ & $[-0.020, +0.017]$ & $-0.06$ & .668 & .668 \\
\bottomrule
\end{tabular*}
\end{threeparttable}
\end{table*}

\subsection{RQ2: model capability}
\label{app:rq2}

RQ2 compares interpretation quality across model configurations, with within-family comparisons used to examine reasoning configuration and model scale. Figure~\ref{fig:RQ2} reports the overall scores, while Tables~\ref{tab:rq2_criteria_ps} and~\ref{tab:rq2_criteria_bll} provide the per-criterion scores for all twelve configurations. Tables~\ref{tab:rq2_contrasts_ps} and~\ref{tab:rq2_contrasts_bll} report the corresponding comparisons on the overall score.

Across both environments, \compl{} and \compr{} are close to ceiling for most models, while \corr{} and especially \just{} show greater variation. In \bl{}, the open-weight thinking models generally have higher \corr{} than \just{}, whereas GPT-5 has similar scores on the two criteria. In \ps{}, \just{} is generally closer to or above \corr{} for the open-weight thinking models. These per-criterion scores provide the basis for the correctness--justifiedness patterns discussed in the main text.

Within Qwen3, thinking configurations outperform their matched instruct configurations at 4B, 30B, and 80B in both environments. Within the Qwen3 thinking models, the effects of scaling are less consistent. The 30B model significantly outperforms 4B in both environments. The 80B model is lower than 30B in \ps{} and does not differ significantly from 30B in \bl{}. GPT-OSS-120B does not significantly outperform GPT-OSS-20B in \ps{}, but does so in \bl{}. The GPT-family comparisons are reported separately because those models differ in training as well as model configuration.

\begin{table*}[tp]
\centering
\begin{threeparttable}
\caption{Per-criterion scores by model, \ps{} ($n = 205$ students).}
\label{tab:rq2_criteria_ps}
\scriptsize
\setlength{\tabcolsep}{3pt}
\begin{tabular*}{\linewidth}{@{\extracolsep{\fill}} l l c c c c c}
\toprule
\textbf{Model} & \textbf{Config.}
  & {\textbf{Complete}} & {\textbf{Correct}} & {\textbf{Justified}} & {\textbf{Compreh.}} & {\textbf{Overall}} \\
\midrule
Qwen3-4B & Instruct & 1.000 & \meanpm{0.512}{0.023} & \meanpm{0.478}{0.020} & \meanpm{0.974}{0.011} & \meanpm{0.385}{0.021} \\
Qwen3-4B & Thinking & 1.000 & \meanpm{0.839}{0.015} & \meanpm{0.924}{0.010} & 1.000 & \meanpm{0.793}{0.017} \\
Qwen3-30B & Instruct & \meanpm{0.972}{0.008} & \meanpm{0.489}{0.023} & \meanpm{0.611}{0.022} & \meanpm{0.964}{0.009} & \meanpm{0.434}{0.022} \\
Qwen3-30B & Thinking & 1.000 & \meanpm{0.896}{0.013} & \meanpm{0.953}{0.008} & \meanpm{0.985}{0.008} & \meanpm{0.846}{0.016} \\
Qwen3-80B & Instruct & 1.000 & \meanpm{0.593}{0.022} & \meanpm{0.660}{0.019} & \meanpm{0.997}{0.003} & \meanpm{0.520}{0.022} \\
Qwen3-80B & Thinking & \meanpm{0.997}{0.003} & \meanpm{0.831}{0.016} & \meanpm{0.922}{0.011} & 1.000 & \meanpm{0.789}{0.018} \\
\midrule
GPT-OSS-20B & Thinking & \meanpm{0.995}{0.005} & \meanpm{0.896}{0.012} & \meanpm{0.946}{0.010} & \meanpm{0.993}{0.005} & \meanpm{0.859}{0.015} \\
GPT-OSS-120B & Thinking & 1.000 & \meanpm{0.928}{0.010} & \meanpm{0.951}{0.009} & 1.000 & \meanpm{0.889}{0.013} \\
\midrule
GPT-4o & Instruct & 1.000 & \meanpm{0.719}{0.018} & \meanpm{0.797}{0.017} & 1.000 & \meanpm{0.667}{0.019} \\
GPT-5-nano & Thinking & 1.000 & \meanpm{0.798}{0.017} & \meanpm{0.595}{0.025} & \meanpm{0.995}{0.005} & \meanpm{0.520}{0.025} \\
GPT-5-mini & Thinking & 1.000 & \meanpm{0.906}{0.012} & \meanpm{0.896}{0.017} & 1.000 & \meanpm{0.829}{0.019} \\
GPT-5 & Thinking & 1.000 & \textbf{\meanpm{0.961}{0.008}} & \textbf{\meanpm{0.984}{0.009}} & 1.000 & \textbf{\meanpm{0.945}{0.011}} \\
\bottomrule
\end{tabular*}
\end{threeparttable}
\end{table*}
\begin{table*}[tp]
\centering
\begin{threeparttable}
\caption{Per-criterion scores by model, \bl{} ($n = 160$ students). GPT-5's \just{} and overall scores use the re-verified exploration counts, as in Table~\ref{tab:rq1}. The other eleven backbones use the stored judge ratings because the claimed counts required for re-verification were recoverable only for GPT-5. Contrasts involving GPT-5 in Table~\ref{tab:rq2_contrasts_bll} use the same re-verified scores.}
\label{tab:rq2_criteria_bll}
\scriptsize
\setlength{\tabcolsep}{3pt}
\begin{tabular*}{\linewidth}{@{\extracolsep{\fill}} l l c c c c c}
\toprule
\textbf{Model} & \textbf{Config.}
  & {\textbf{Complete}} & {\textbf{Correct}} & {\textbf{Justified}} & {\textbf{Compreh.}} & {\textbf{Overall}} \\
\midrule
Qwen3-4B & Instruct & \meanpm{0.956}{0.008} & \meanpm{0.596}{0.016} & \meanpm{0.263}{0.015} & \meanpm{0.987}{0.007} & \meanpm{0.147}{0.013} \\
Qwen3-4B & Thinking & \meanpm{0.988}{0.005} & \meanpm{0.690}{0.016} & \meanpm{0.410}{0.017} & 1.000 & \meanpm{0.335}{0.017} \\
Qwen3-30B & Instruct & \meanpm{0.890}{0.014} & \meanpm{0.642}{0.015} & \meanpm{0.307}{0.015} & \meanpm{0.966}{0.012} & \meanpm{0.182}{0.015} \\
Qwen3-30B & Thinking & \meanpm{0.983}{0.004} & \meanpm{0.791}{0.013} & \meanpm{0.478}{0.018} & \meanpm{0.988}{0.009} & \meanpm{0.408}{0.018} \\
Qwen3-80B & Instruct & \meanpm{0.989}{0.004} & \meanpm{0.615}{0.016} & \meanpm{0.337}{0.018} & 1.000 & \meanpm{0.224}{0.018} \\
Qwen3-80B & Thinking & \meanpm{0.999}{0.001} & \meanpm{0.812}{0.012} & \meanpm{0.512}{0.020} & 1.000 & \meanpm{0.451}{0.019} \\
\midrule
GPT-OSS-20B & Thinking & \meanpm{0.999}{0.001} & \meanpm{0.731}{0.015} & \meanpm{0.653}{0.018} & 1.000 & \meanpm{0.519}{0.020} \\
GPT-OSS-120B & Thinking & 1.000 & \meanpm{0.767}{0.012} & \meanpm{0.688}{0.017} & 1.000 & \meanpm{0.574}{0.016} \\
\midrule
GPT-4o & Instruct & 1.000 & \meanpm{0.588}{0.015} & \meanpm{0.486}{0.018} & 1.000 & \meanpm{0.374}{0.017} \\
GPT-5-nano & Thinking & \meanpm{0.930}{0.012} & \meanpm{0.609}{0.017} & \meanpm{0.281}{0.016} & 1.000 & \meanpm{0.226}{0.016} \\
GPT-5-mini & Thinking & 1.000 & \meanpm{0.769}{0.013} & \meanpm{0.774}{0.017} & 1.000 & \meanpm{0.650}{0.017} \\
GPT-5 & Thinking & 1.000 & \meanpm{0.815}{0.012} & \textbf{\meanpm{0.833}{0.014}} & 1.000 & \textbf{\meanpm{0.726}{0.015}} \\
\bottomrule
\end{tabular*}
\end{threeparttable}
\end{table*}
\begin{table*}[tp]
\centering
\begin{threeparttable}
\caption{RQ2 contrasts on the overall score in PharmaSim. Comparisons are paired at the student level with 95\% cluster-bootstrap intervals. Tests use the Wilcoxon signed-rank test. Bold $p_{\text{Holm}}$ marks $p < .05$.}
\label{tab:rq2_contrasts_ps}
\footnotesize
\setlength{\tabcolsep}{3pt}
\begin{tabular*}{\linewidth}{@{\extracolsep{\fill}} l l c c c l}
\toprule
\textbf{Contrast} & $\Delta$ \textbf{[95\% CI]} & $r_{\text{rb}}$
  & \textbf{$p$} & \textbf{$p_{\text{Holm}}$} & \textbf{Correction family} \\
\midrule
Qwen3-4B Thinking $-$ Instruct & $+0.408$ $[+0.354, +0.463]$ & $+0.86$ & $<$.001 & $\mathbf{<}$\textbf{.001} & thinking vs.\ instruct ($k=6$) \\
Qwen3-30B Thinking $-$ Instruct & $+0.411$ $[+0.358, +0.465]$ & $+0.92$ & $<$.001 & $\mathbf{<}$\textbf{.001} & thinking vs.\ instruct ($k=6$) \\
Qwen3-80B Thinking $-$ Instruct & $+0.268$ $[+0.216, +0.319]$ & $+0.76$ & $<$.001 & $\mathbf{<}$\textbf{.001} & thinking vs.\ instruct ($k=6$) \\
Qwen3 30B-Thinking $-$ 4B-Thinking & $+0.052$ $[+0.013, +0.091]$ & $+0.32$ & .005 & \textbf{.012} & Qwen3 scale ($k=3$) \\
Qwen3 80B-Thinking $-$ 30B-Thinking & $-0.057$ $[-0.098, -0.018]$ & $-0.33$ & .004 & \textbf{.012} & Qwen3 scale ($k=3$) \\
Qwen3 80B-Thinking $-$ 4B-Thinking & $-0.005$ $[-0.046, +0.034]$ & $+0.02$ & .883 & .883 & Qwen3 scale ($k=3$) \\
GPT-5-nano $-$ GPT-4o & $-0.146$ $[-0.203, -0.088]$ & $-0.39$ & $<$.001 & $\mathbf{<}$\textbf{.001} & GPT pairs ($k=6$) \\
GPT-5-mini $-$ GPT-4o & $+0.163$ $[+0.114, +0.211]$ & $+0.65$ & $<$.001 & $\mathbf{<}$\textbf{.001} & GPT pairs ($k=6$) \\
GPT-5 $-$ GPT-4o & $+0.278$ $[+0.236, +0.322]$ & $+0.89$ & $<$.001 & $\mathbf{<}$\textbf{.001} & GPT pairs ($k=6$) \\
GPT-5-mini $-$ GPT-5-nano & $+0.309$ $[+0.254, +0.363]$ & $+0.83$ & $<$.001 & $\mathbf{<}$\textbf{.001} & GPT pairs ($k=6$) \\
GPT-5 $-$ GPT-5-nano & $+0.424$ $[+0.371, +0.475]$ & $+0.94$ & $<$.001 & $\mathbf{<}$\textbf{.001} & GPT pairs ($k=6$) \\
GPT-5 $-$ GPT-5-mini & $+0.115$ $[+0.075, +0.156]$ & $+0.71$ & $<$.001 & $\mathbf{<}$\textbf{.001} & GPT pairs ($k=6$) \\
GPT-5 $-$ GPT-OSS-120B & $+0.055$ $[+0.026, +0.085]$ & $+0.56$ & $<$.001 & $\mathbf{<}$\textbf{.001} & single, uncorrected \\
mean(Think-Instr) $-$ (80B-Instr $-$ 4B-Instr) & $+0.228$ $[+0.170, +0.285]$ & {--} &  {--}  &  {--}  & difference of contrasts \\
GPT-OSS 120B $-$ 20B & $+0.031$ $[-0.007, +0.068]$ & $+0.22$ & .068 & .068 & single, uncorrected \\
\bottomrule
\end{tabular*}
\begin{tablenotes}[flushleft]\footnotesize
\item $r_{\text{rb}}$ is the matched-pairs rank-biserial correlation. A dash marks a
      quantity that does not apply: the reasoning-gap contrast is a difference of
      two observed contrasts, assessed by its bootstrap interval rather than by a
      signed-rank test.
\item $\Delta$ and $r_{\text{rb}}$ can have opposite signs when the mean difference is close to zero, although this does not occur here. They summarize paired differences differently: $\Delta$ is the mean of the raw paired differences, whereas $r_{\text{rb}}$ reflects the relative signed ranks of their magnitudes. As a result, many smaller differences in one direction can outweigh fewer larger differences in the other direction in $r_{\text{rb}}$, even when $\Delta$ points the other way.

\item Correction families. \emph{Thinking vs.\ instruct} is corrected across the
      three Qwen3 sizes in \emph{both} environments, per criterion ($k=6$), so three of
      its six members sit in the companion table rather than here. \emph{Qwen3 scale}
      and \emph{GPT pairs} are corrected within this environment and criterion ($k=3$
      and $k=6$), and all their members are shown. Rows marked \emph{single} take no
      correction.
\end{tablenotes}
\end{threeparttable}
\end{table*}
\begin{table*}[tp]
\centering
\begin{threeparttable}
\caption{RQ2 contrasts on the overall score in Beer's Law Lab. Comparisons are paired at the student level with 95\% cluster-bootstrap intervals. Tests use the Wilcoxon signed-rank test. Bold $p_{\text{Holm}}$ marks $p < .05$.}
\label{tab:rq2_contrasts_bll}
\footnotesize
\setlength{\tabcolsep}{3pt}
\begin{tabular*}{\linewidth}{@{\extracolsep{\fill}} l l c c c l}
\toprule
\textbf{Contrast} & $\Delta$ \textbf{[95\% CI]} & $r_{\text{rb}}$
  & \textbf{$p$} & \textbf{$p_{\text{Holm}}$} & \textbf{Correction family} \\
\midrule
Qwen3-4B Thinking $-$ Instruct & $+0.189$ $[+0.153, +0.225]$ & $+0.77$ & $<$.001 & $\mathbf{<}$\textbf{.001} & thinking vs.\ instruct ($k=6$) \\
Qwen3-30B Thinking $-$ Instruct & $+0.226$ $[+0.190, +0.263]$ & $+0.89$ & $<$.001 & $\mathbf{<}$\textbf{.001} & thinking vs.\ instruct ($k=6$) \\
Qwen3-80B Thinking $-$ Instruct & $+0.227$ $[+0.185, +0.268]$ & $+0.81$ & $<$.001 & $\mathbf{<}$\textbf{.001} & thinking vs.\ instruct ($k=6$) \\
Qwen3 30B-Thinking $-$ 4B-Thinking & $+0.072$ $[+0.033, +0.110]$ & $+0.39$ & $<$.001 & $\mathbf{<}$\textbf{.001} & Qwen3 scale ($k=3$) \\
Qwen3 80B-Thinking $-$ 30B-Thinking & $+0.043$ $[+0.005, +0.082]$ & $+0.19$ & .066 & .066 & Qwen3 scale ($k=3$) \\
Qwen3 80B-Thinking $-$ 4B-Thinking & $+0.115$ $[+0.078, +0.153]$ & $+0.55$ & $<$.001 & $\mathbf{<}$\textbf{.001} & Qwen3 scale ($k=3$) \\
GPT-5-nano $-$ GPT-4o & $-0.148$ $[-0.184, -0.113]$ & $-0.67$ & $<$.001 & $\mathbf{<}$\textbf{.001} & GPT pairs ($k=6$) \\
GPT-5-mini $-$ GPT-4o & $+0.276$ $[+0.242, +0.312]$ & $+0.96$ & $<$.001 & $\mathbf{<}$\textbf{.001} & GPT pairs ($k=6$) \\
GPT-5 $-$ GPT-4o & $+0.353$ $[+0.320, +0.385]$ & $+0.99$ & $<$.001 & $\mathbf{<}$\textbf{.001} & GPT pairs ($k=6$) \\
GPT-5-mini $-$ GPT-5-nano & $+0.424$ $[+0.388, +0.461]$ & $+1.00$ & $<$.001 & $\mathbf{<}$\textbf{.001} & GPT pairs ($k=6$) \\
GPT-5 $-$ GPT-5-nano & $+0.501$ $[+0.464, +0.537]$ & $+1.00$ & $<$.001 & $\mathbf{<}$\textbf{.001} & GPT pairs ($k=6$) \\
GPT-5 $-$ GPT-5-mini & $+0.076$ $[+0.050, +0.103]$ & $+0.55$ & $<$.001 & $\mathbf{<}$\textbf{.001} & GPT pairs ($k=6$) \\
GPT-5 $-$ GPT-OSS-120B & $+0.153$ $[+0.124, +0.182]$ & $+0.86$ & $<$.001 & $\mathbf{<}$\textbf{.001} & single, uncorrected \\
mean(Think-Instr) $-$ (80B-Instr $-$ 4B-Instr) & $+0.137$ $[+0.093, +0.178]$ & {--} &  {--}  &  {--}  & difference of contrasts \\
GPT-OSS 120B $-$ 20B & $+0.055$ $[+0.018, +0.091]$ & $+0.32$ & .003 & \textbf{.003} & single, uncorrected \\
\bottomrule
\end{tabular*}
\begin{tablenotes}[flushleft]\footnotesize
\item $r_{\text{rb}}$ is the matched-pairs rank-biserial correlation. A dash marks a
      quantity that does not apply: the reasoning-gap contrast is a difference of
      two observed contrasts, assessed by its bootstrap interval rather than by a
      signed-rank test.
\item $\Delta$ and $r_{\text{rb}}$ can have opposite signs when the mean difference is close to zero, although this does not occur here. They summarize paired differences differently: $\Delta$ is the mean of the raw paired differences, whereas $r_{\text{rb}}$ reflects the relative signed ranks of their magnitudes. As a result, many smaller differences in one direction can outweigh fewer larger differences in the other direction in $r_{\text{rb}}$, even when $\Delta$ points the other way.

\item Correction families. \emph{Thinking vs.\ instruct} is corrected across the
      three Qwen3 sizes in \emph{both} environments, per criterion ($k=6$), so three of
      its six members sit in the companion table rather than here. \emph{Qwen3 scale}
      and \emph{GPT pairs} are corrected within this environment and criterion ($k=3$
      and $k=6$), and all their members are shown. Rows marked \emph{single} take no
      correction.
\end{tablenotes}
\end{threeparttable}
\end{table*}

\subsection{RQ3: external guidance}
\label{app:rq3}

RQ3 compares four forms of external guidance with the same model's no-guidance baseline. Expert-authored guidance consists of additional instructions and examples, while clickstream-extracted guidance consists of action indices and action counts. Tables~\ref{tab:rq3_criteria_ps} and~\ref{tab:rq3_criteria_bll} report the per-criterion scores, and Tables~\ref{tab:rq3_contrasts_ps} and~\ref{tab:rq3_contrasts_bll} report comparisons on the overall score.

For each model and environment, the four guidance-versus-baseline contrasts form one Holm correction family ($k=4$). The comparison between clickstream-extracted and expert-authored guidance is corrected jointly across two models, both environments, and three outcomes ($k=12$). As elsewhere, $\Delta$ reports the paired mean difference and $r_{\text{rb}}$ the matched-pairs rank-biserial correlation. Because the two summarize paired differences differently, their signs need not agree when the mean difference is close to zero.

\begin{table*}[tp]
\centering
\begin{threeparttable}
\caption{Per-criterion scores by guidance condition, \ps{} ($n = 205$ students).}
\label{tab:rq3_criteria_ps}
\scriptsize
\setlength{\tabcolsep}{3pt}
\begin{tabular*}{\linewidth}{@{\extracolsep{\fill}} l l c c c c c}
\toprule
\textbf{Model} & \textbf{Guidance}
  & {\textbf{Complete}} & {\textbf{Correct}} & {\textbf{Justified}} & {\textbf{Compreh.}} & {\textbf{Overall}} \\
\midrule
Qwen3-80B-Thinking & None (baseline) & \meanpm{0.998}{0.002} & \meanpm{0.849}{0.016} & \meanpm{0.868}{0.013} & 1.000 & \meanpm{0.756}{0.019} \\
Qwen3-80B-Thinking & Instructions & 1.000 & \meanpm{0.836}{0.017} & \meanpm{0.894}{0.013} & 1.000 & \meanpm{0.763}{0.020} \\
Qwen3-80B-Thinking & Examples & \meanpm{0.995}{0.005} & \meanpm{0.828}{0.016} & \meanpm{0.891}{0.013} & \meanpm{0.995}{0.005} & \meanpm{0.751}{0.018} \\
Qwen3-80B-Thinking & Action indices & 1.000 & \meanpm{0.859}{0.016} & \meanpm{0.904}{0.012} & 1.000 & \meanpm{0.793}{0.018} \\
Qwen3-80B-Thinking & Action counts & 1.000 & \meanpm{0.834}{0.016} & \meanpm{0.885}{0.014} & 1.000 & \meanpm{0.758}{0.018} \\
\midrule
GPT-OSS-120B & None (baseline) & 1.000 & \meanpm{0.928}{0.010} & \meanpm{0.951}{0.009} & 1.000 & \meanpm{0.889}{0.013} \\
GPT-OSS-120B & Instructions & 1.000 & \meanpm{0.919}{0.012} & \meanpm{0.956}{0.009} & 1.000 & \meanpm{0.888}{0.014} \\
GPT-OSS-120B & Examples & 1.000 & \meanpm{0.919}{0.011} & \meanpm{0.956}{0.009} & \meanpm{0.995}{0.005} & \meanpm{0.891}{0.013} \\
GPT-OSS-120B & Action indices & 1.000 & \meanpm{0.906}{0.011} & \meanpm{0.958}{0.008} & \meanpm{0.995}{0.005} & \meanpm{0.872}{0.013} \\
GPT-OSS-120B & Action counts & 1.000 & \meanpm{0.906}{0.011} & \meanpm{0.927}{0.011} & \meanpm{0.990}{0.007} & \meanpm{0.846}{0.014} \\
\bottomrule
\end{tabular*}
\end{threeparttable}
\end{table*}
\begin{table*}[tp]
\centering
\begin{threeparttable}
\caption{Per-criterion scores by guidance condition, \bl{} ($n = 160$ students).}
\label{tab:rq3_criteria_bll}
\scriptsize
\setlength{\tabcolsep}{3pt}
\begin{tabular*}{\linewidth}{@{\extracolsep{\fill}} l l c c c c c}
\toprule
\textbf{Model} & \textbf{Guidance}
  & {\textbf{Complete}} & {\textbf{Correct}} & {\textbf{Justified}} & {\textbf{Compreh.}} & {\textbf{Overall}} \\
\midrule
Qwen3-80B-Thinking & None (baseline) & \meanpm{0.999}{0.001} & \meanpm{0.812}{0.012} & \meanpm{0.512}{0.020} & 1.000 & \meanpm{0.451}{0.019} \\
Qwen3-80B-Thinking & Instructions & \meanpm{0.996}{0.003} & \meanpm{0.811}{0.014} & \meanpm{0.564}{0.019} & 1.000 & \meanpm{0.476}{0.019} \\
Qwen3-80B-Thinking & Examples & \meanpm{0.998}{0.002} & \meanpm{0.838}{0.011} & \meanpm{0.600}{0.020} & 1.000 & \meanpm{0.517}{0.019} \\
Qwen3-80B-Thinking & Action indices & \meanpm{0.997}{0.002} & \meanpm{0.800}{0.012} & \meanpm{0.735}{0.017} & 1.000 & \meanpm{0.601}{0.018} \\
Qwen3-80B-Thinking & Action counts & 1.000 & \meanpm{0.819}{0.012} & \meanpm{0.830}{0.013} & 1.000 & \meanpm{0.703}{0.015} \\
\midrule
GPT-OSS-120B & None (baseline) & 1.000 & \meanpm{0.767}{0.012} & \meanpm{0.688}{0.017} & 1.000 & \meanpm{0.574}{0.016} \\
GPT-OSS-120B & Instructions & 1.000 & \meanpm{0.749}{0.012} & \meanpm{0.701}{0.016} & 1.000 & \meanpm{0.558}{0.016} \\
GPT-OSS-120B & Examples & 1.000 & \meanpm{0.760}{0.012} & \meanpm{0.713}{0.017} & 1.000 & \meanpm{0.585}{0.016} \\
GPT-OSS-120B & Action indices & 1.000 & \meanpm{0.758}{0.013} & \meanpm{0.843}{0.012} & 1.000 & \meanpm{0.677}{0.014} \\
GPT-OSS-120B & Action counts & 1.000 & \meanpm{0.765}{0.013} & \meanpm{0.874}{0.010} & 1.000 & \meanpm{0.702}{0.014} \\
\bottomrule
\end{tabular*}
\end{threeparttable}
\end{table*}
\begin{table*}[tp]
\centering
\begin{threeparttable}
\caption{RQ3 contrasts on the overall score in PharmaSim. Each guidance condition is compared with the same model's no-guidance baseline. The pooled clickstream-extracted condition combines action indices and action counts, and the pooled expert-authored condition combines instructions and examples. $\Delta$ is the paired mean difference and $r_{\text{rb}}$ the matched-pairs rank-biserial correlation. Bold $p_{\text{Holm}}$ marks $p < .05$.}
\label{tab:rq3_contrasts_ps}
\footnotesize
\setlength{\tabcolsep}{3pt}
\begin{tabular*}{\linewidth}{@{\extracolsep{\fill}} l l c c c l}
\toprule
\textbf{Contrast} & $\Delta$ \textbf{[95\% CI]} & $r_{\text{rb}}$
  & \textbf{$p$} & \textbf{$p_{\text{Holm}}$} & \textbf{Holm family} \\
\midrule
GPT-OSS-120B: instructions $-$ baseline & $-0.002$ $[-0.031, +0.028]$ & $+0.03$ & .836 & $>$.999 & baseline ($k=4$) \\
GPT-OSS-120B: examples $-$ baseline & $+0.002$ $[-0.028, +0.031]$ & $+0.01$ & .965 & $>$.999 & baseline ($k=4$) \\
GPT-OSS-120B: action indices $-$ baseline & $-0.018$ $[-0.047, +0.010]$ & $-0.22$ & .119 & .358 & baseline ($k=4$) \\
GPT-OSS-120B: action counts $-$ baseline & $-0.044$ $[-0.075, -0.015]$ & $-0.47$ & $<$.001 & \textbf{.001} & baseline ($k=4$) \\
Qwen3-80B-Thinking: instructions $-$ baseline & $+0.007$ $[-0.036, +0.049]$ & $+0.06$ & .606 & $>$.999 & baseline ($k=4$) \\
Qwen3-80B-Thinking: examples $-$ baseline & $-0.005$ $[-0.047, +0.037]$ & $-0.08$ & .476 & $>$.999 & baseline ($k=4$) \\
Qwen3-80B-Thinking: action indices $-$ baseline & $+0.037$ $[-0.003, +0.076]$ & $+0.19$ & .097 & .386 & baseline ($k=4$) \\
Qwen3-80B-Thinking: action counts $-$ baseline & $+0.002$ $[-0.042, +0.046]$ & $-0.02$ & .855 & $>$.999 & baseline ($k=4$) \\
GPT-OSS-120B: extracted $-$ expert & $-0.031$ $[-0.054, -0.008]$ & $-0.30$ & .006 & \textbf{.048} & guidance type ($k=12$) \\
Qwen3-80B-Thinking: extracted $-$ expert & $+0.019$ $[-0.012, +0.050]$ & $+0.10$ & .340 & .881 & guidance type ($k=12$) \\
\bottomrule
\end{tabular*}
\end{threeparttable}
\end{table*}
\begin{table*}[tp]
\centering
\begin{threeparttable}
\caption{RQ3 contrasts on the overall score in Beer's Law Lab. Each guidance condition is compared with the same model's no-guidance baseline. The pooled clickstream-extracted condition combines action indices and action counts, and the pooled expert-authored condition combines instructions and examples. $\Delta$ is the paired mean difference and $r_{\text{rb}}$ the matched-pairs rank-biserial correlation. Bold $p_{\text{Holm}}$ marks $p < .05$.}
\label{tab:rq3_contrasts_bll}
\footnotesize
\setlength{\tabcolsep}{3pt}
\begin{tabular*}{\linewidth}{@{\extracolsep{\fill}} l l c c c l}
\toprule
\textbf{Contrast} & $\Delta$ \textbf{[95\% CI]} & $r_{\text{rb}}$
  & \textbf{$p$} & \textbf{$p_{\text{Holm}}$} & \textbf{Holm family} \\
\midrule
GPT-OSS-120B: instructions $-$ baseline & $-0.015$ $[-0.041, +0.009]$ & $-0.19$ & .084 & .167 & baseline ($k=4$) \\
GPT-OSS-120B: examples $-$ baseline & $+0.012$ $[-0.018, +0.042]$ & $+0.02$ & .837 & .837 & baseline ($k=4$) \\
GPT-OSS-120B: action indices $-$ baseline & $+0.103$ $[+0.074, +0.133]$ & $+0.62$ & $<$.001 & $\mathbf{<}$\textbf{.001} & baseline ($k=4$) \\
GPT-OSS-120B: action counts $-$ baseline & $+0.128$ $[+0.098, +0.158]$ & $+0.77$ & $<$.001 & $\mathbf{<}$\textbf{.001} & baseline ($k=4$) \\
Qwen3-80B-Thinking: instructions $-$ baseline & $+0.026$ $[-0.012, +0.063]$ & $+0.17$ & .086 & .086 & baseline ($k=4$) \\
Qwen3-80B-Thinking: examples $-$ baseline & $+0.066$ $[+0.029, +0.103]$ & $+0.31$ & .003 & \textbf{.005} & baseline ($k=4$) \\
Qwen3-80B-Thinking: action indices $-$ baseline & $+0.150$ $[+0.112, +0.188]$ & $+0.65$ & $<$.001 & $\mathbf{<}$\textbf{.001} & baseline ($k=4$) \\
Qwen3-80B-Thinking: action counts $-$ baseline & $+0.253$ $[+0.215, +0.291]$ & $+0.88$ & $<$.001 & $\mathbf{<}$\textbf{.001} & baseline ($k=4$) \\
GPT-OSS-120B: extracted $-$ expert & $+0.118$ $[+0.096, +0.139]$ & $+0.85$ & $<$.001 & $\mathbf{<}$\textbf{.001} & guidance type ($k=12$) \\
Qwen3-80B-Thinking: extracted $-$ expert & $+0.156$ $[+0.128, +0.183]$ & $+0.81$ & $<$.001 & $\mathbf{<}$\textbf{.001} & guidance type ($k=12$) \\
\bottomrule
\end{tabular*}
\end{threeparttable}
\end{table*}

\section{Per-strategy analysis}
\label{app:per_strategy}
\begin{figure*}[tp]
\centering
\includegraphics[width=0.8\textwidth]{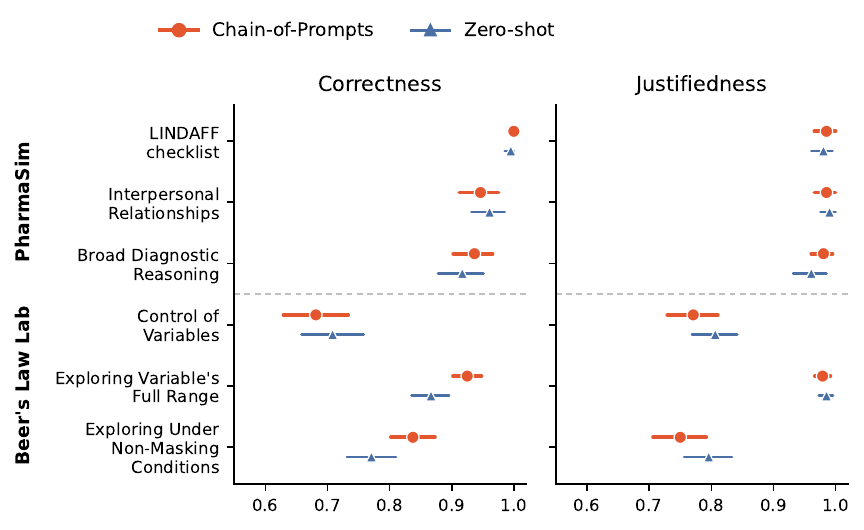}
\caption{Per-strategy \corr{} and \just{} under \cop{} and \zs{}, with 95\,\% cluster-bootstrap confidence intervals over students. Performance is shown separately for the three strategies in each environment.}
\label{fig:per_strategy}
\end{figure*}

The main results average scores across strategies. Figure~\ref{fig:per_strategy} therefore reports \corr{} and \just{} separately for each strategy under GPT-5 with \cop{} and \zs{}. We focus on these two prompting methods because \ct{} does not significantly outperform either method on the overall score in RQ1.

All three \bl{} strategies have four adherence levels. \ps{}'s \emph{Interpersonal Relationships} strategy also has four levels but does not depend on the order or timing of actions. Its \corr{} is 0.95--0.96, compared with 1.00 for the binary \emph{LINDAFF checklist}, 0.92--0.94 for the binary \emph{Broad Diagnostic Reasoning}, and 0.68--0.93 across the three \bl{} strategies. In these data, the number of adherence levels alone therefore does not account for the lower scores in \bl{}.

Performance also differs among the \bl{} strategies. \emph{Control of Variables}, whose justification criteria require exact exploration counts, has the lowest \corr{} under both \cop{} and \zs{} at 0.68 and 0.71. \emph{Exploring a Variable's Full Range} has the highest \corr{} at 0.93 and 0.87, and the highest \just{} at 0.98 and 0.99. These differences are consistent with the greater demands that some \bl{} strategies place on reasoning over action order, timing, and exact counts.

\section{Exploratory analysis of clickstream length}
\label{app:length}

We conducted an exploratory analysis to examine whether interpretation quality is associated with the length of the formatted clickstream and whether external guidance changes this relationship. The analysis was developed after inspecting the RQ3 results and was not planned in advance. It reuses the two models, guidance conditions, and \cop{} prompting method from RQ3. The results should therefore be interpreted as descriptive rather than confirmatory.

We measure input length in tokens because the model receives the formatted clickstream as text. Although the two environments overlap substantially in the number of logged events, their formatted traces differ considerably in token length. The median formatted trace contains 9{,}568 tokens in \bl{} and 1{,}108 in \ps{}, largely because a formatted \bl{} event contains more information (Table~\ref{tab:length_descriptives}).

\begin{table*}[tp]
\centering
\begin{threeparttable}
\caption{Clickstream length by environment. Event count is the number of logged actions in a session. Tokens per event is the formatted token count divided by the event count. Formatted token count is the length of the exact clickstream string supplied to the model using the \texttt{o200k\_base} tokenizer. Skew is sample skewness.}
\label{tab:length_descriptives}
\small
\setlength{\tabcolsep}{4pt}
\begin{tabular*}{\linewidth}{@{\extracolsep{\fill}}l l r r r r r r}
\toprule
\textbf{Environment} & \textbf{Metric} & \textbf{$n$} & \textbf{Median} & \textbf{IQR} & \textbf{Min} & \textbf{Max} & \textbf{Skew} \\
\midrule
\multirow{3}{*}{\bl{}}
  & Event count            & 160 & 38    & 32.5 & 5   & 217   & 1.95 \\
  & Tokens per event        & 160 & 229   & 77.2 & 119 & 555   & 1.37 \\
  & Formatted token count   & 160 & 9{,}568 & 8{,}250 & 889 & 39{,}241 & 1.43 \\
\midrule
\multirow{3}{*}{\ps{}}
  & Event count            & 205 & 22    & 15   & 1   & 59    & 0.38 \\
  & Tokens per event        & 205 & 51    & 2.6  & 44  & 78    & 3.77 \\
  & Formatted token count   & 205 & 1{,}108 & 745   & 50  & 2{,}817  & 0.37 \\
\bottomrule
\end{tabular*}
\end{threeparttable}
\end{table*}

\begin{figure*}[tp]
\centering
\includegraphics[width=0.8\textwidth]{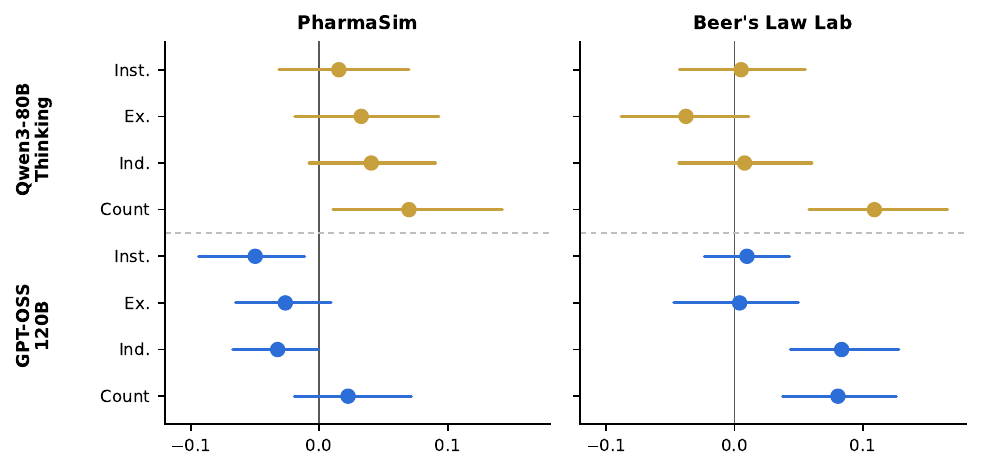}
\caption{Difference in the slope relating overall score to $\log$(formatted token count) between each guidance condition and the corresponding no-guidance baseline. Positive values indicate a weaker decline with trace length under guidance. Error bars show 95\,\% cluster-bootstrap confidence intervals. The intervals are descriptive and no significance is claimed.}
\label{fig:rq3_length}
\end{figure*}

Overall score decreases with $\log$(formatted token count) in both environments. Figure~\ref{fig:rq3_length} compares the slope under each guidance condition with the no-guidance slope for the same model. Because this analysis was selected after observing the RQ3 results, we report the 16 slope differences with 95\,\% cluster-bootstrap confidence intervals as descriptive effect sizes and do not perform significance tests. The three largest positive differences occur in \bl{} under clickstream-extracted guidance: $+0.109$ $[+0.058, +0.166]$ for action counts with Qwen3-80B-Thinking, $+0.083$ $[+0.043, +0.128]$ for action indices with GPT-OSS-120B, and $+0.080$ $[+0.037, +0.126]$ for action counts with GPT-OSS-120B. Changes under expert-authored guidance in \bl{} are below $0.04$, and the \ps{} differences are smaller and vary in direction.

We also examined whether differences in formatted trace length could account for the lower GPT-5 scores in \bl{}. The environments overlap little in token count, so this comparison is necessarily limited. After coarsened exact matching on token-count deciles, the difference between environments becomes smaller, but only 14 \bl{} students remain in the matched sample (Table~\ref{tab:env_gap}). As a second check, scores for the most token-heavy third of \ps{} remains close to the rest of \ps{} and above \bl{}, although even these traces are shorter than the median \bl{} trace. Taken together, these analyses indicate that input length may contribute to the environment difference but does not isolate its cause.

\begin{table*}[tp]
\centering
\begin{threeparttable}
\caption{Exploratory comparison of \corr{} and \just{} across environments after accounting for formatted token count. Results use GPT-5 with the three prompting methods pooled. Cells report means with standard errors over students. The token-matched comparison retains only students in token-count deciles represented in both environments.}
\label{tab:env_gap}
\small
\setlength{\tabcolsep}{4pt}
\begin{tabular*}{\linewidth}{@{\extracolsep{\fill}}l l c c}
\toprule
\textbf{Comparison} & \textbf{Environment} & \textbf{\corr{}} & \textbf{\just{}} \\
\midrule
Unrestricted & \bl{} ($n{=}4{,}320$ ratings, 160 students) & \meanpm{0.798}{0.011} & \meanpm{0.845}{0.012} \\
             & \ps{} ($n{=}1{,}845$ ratings, 205 students)          & \meanpm{0.959}{0.006} & \meanpm{0.975}{0.006} \\
\midrule
Token-matched\tnote{a} & \bl{} ($n{=}378$ ratings, 14 students)   & \meanpm{0.905}{0.025} & \meanpm{0.966}{0.011} \\
                        & \ps{} ($n{=}846$ ratings, 94 students)   & \meanpm{0.959}{0.008} & \meanpm{0.974}{0.008} \\
\midrule
\ps{} heaviest tercile & long ($n{=}621$ ratings, 69 students)  & \meanpm{0.940}{0.013} & \meanpm{0.952}{0.014} \\
                       & rest ($n{=}1{,}224$ ratings, 136 students)  & \meanpm{0.969}{0.006} & \meanpm{0.987}{0.005} \\
\bottomrule
\end{tabular*}
\begin{tablenotes}[flushleft]\footnotesize
\item[a] Coarsened exact matching on formatted-token-count decile using pooled bins across both environments. Students are retained only in bins represented in both environments.
\end{tablenotes}
\end{threeparttable}
\end{table*}

\FloatBarrier

\section{Qualitative analysis of selected interpretations}
\label{app:ex_catalogue}

This appendix provides selected model outputs to illustrate the rubric and the differences observed across experimental conditions. Model outputs are reproduced as generated. The marker \texttt{[...]} indicates omitted text, and highlighting is added only for readability. Examples described as randomly sampled were selected with a fixed-seed random-number generator retained with the analysis materials. Full clickstream traces for these cases are provided in the GitHub repository.

\subsection{What a high-quality interpretation looks like}
\label{app:ex_d}

\begin{figure*}[tp]
\centering
\includegraphics[width=\textwidth]{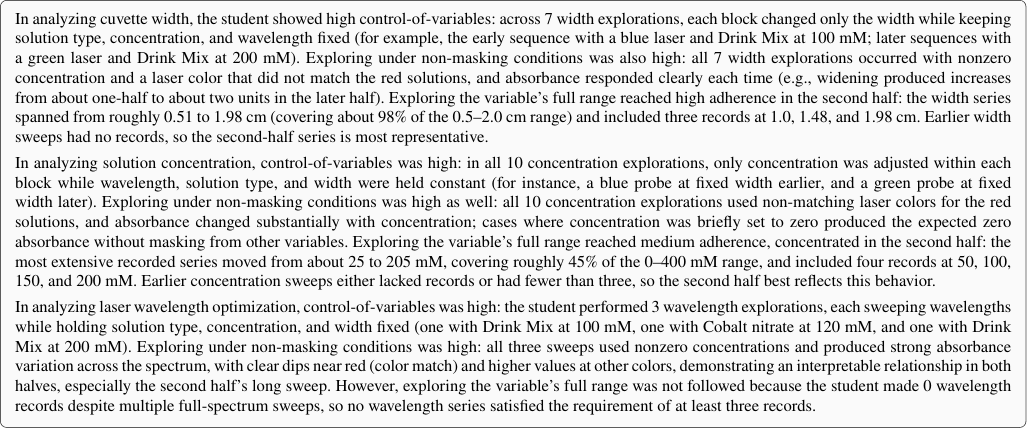}
\caption{Example \bl{} interpretation with GPT-5, \cop{}, no guidance, quoted in full and satisfying every rubric criterion.}
\label{fig:app_c_bll_full}
\end{figure*}

\begin{figure*}[tp]
\centering
\includegraphics[width=\textwidth]{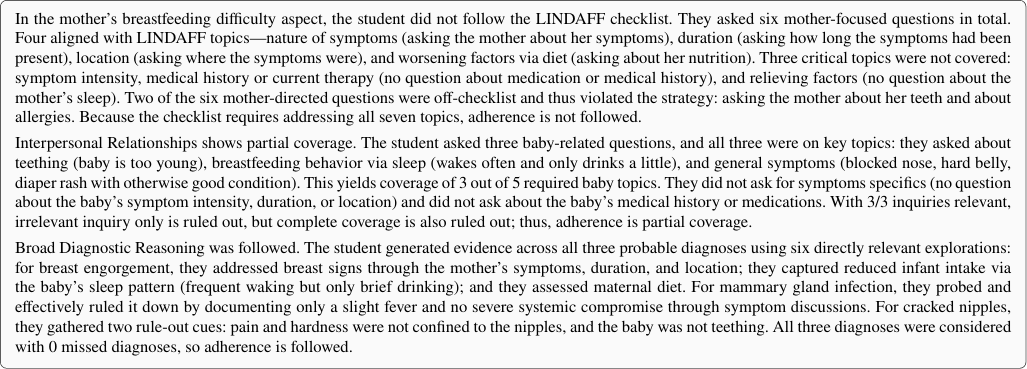}
\caption{Example \ps{} interpretation (GPT-5, \cop{}, no guidance), quoted in full and satisfying every rubric criterion.}
\label{fig:app_c_ps_full}
\end{figure*}

Figures~\ref{fig:app_c_bll_full} and~\ref{fig:app_c_ps_full} show one complete GPT-5 interpretation from each environment under \cop{} with no guidance. Both receive a score of 1 on every criterion for every strategy--aspect pair.

The examples show how the rubric is applied to successful interpretations. Each adherence judgment is supported with strategy-relevant evidence, and each strategy or aspect is addressed separately. The \bl{} example includes high, medium, and not-followed judgments across its nine strategy--aspect pairs.

\subsection{\cop{} and the prompting baselines (RQ1)}
\label{app:ex_c}

\begin{figure*}[tp]
\centering
\includegraphics[width=\textwidth]{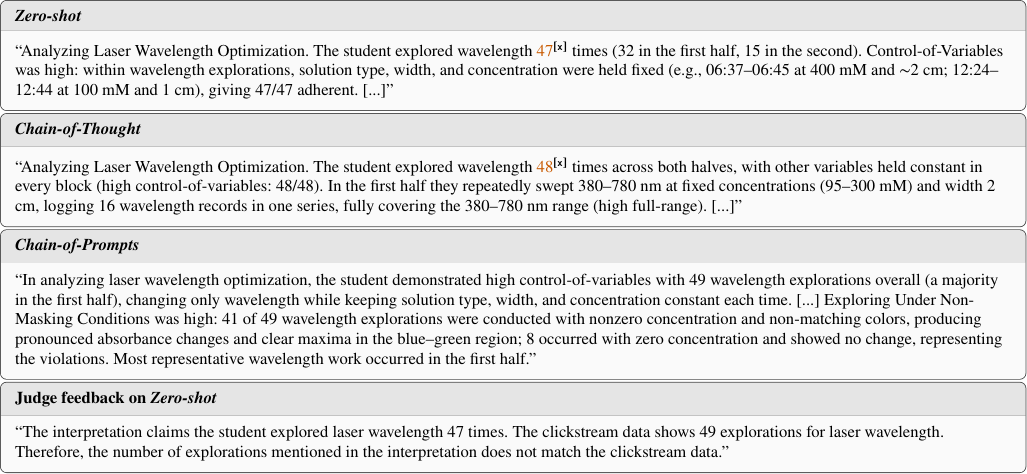}
\caption{The same Control-of-Variables pair for Laser Wavelength Optimization under the three prompting methods with GPT-5 Thinking. The ground-truth label is high, with 49 explorations and 7 recorded violations. All three methods assign the correct level, while \zs{} and \ct{} report 47 and 48 explorations and \cop{} reports 49.}
\label{fig:app_c_methods}
\end{figure*}

Figure~\ref{fig:app_c_methods} compares the three prompting methods on the same strategy--aspect pair. The example was selected from cases in which all three outputs contained a parseable exploration count. All three methods assign the correct high adherence level. They differ in the supporting counts: \zs{} reports 47 explorations, \ct{} reports 48, and \cop{} reports the recorded count of 49. The two baseline outputs also describe all of their counted explorations as compliant, despite seven recorded violations in the ground-truth label. In this example, the difference between methods is therefore in the factual evidence supporting the shared adherence judgment rather than in the judgment itself.

\subsection{Matched instruct and thinking outputs (RQ2)}
\label{app:ex_config}

\begin{figure*}[tp]
\centering
\includegraphics[width=\textwidth]{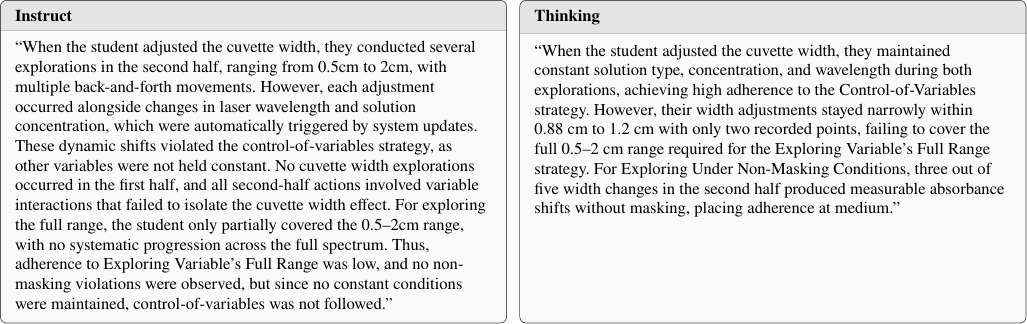}
\caption{The same \bl{} Cuvette Width aspect under \cop{} with no guidance, interpreted by Qwen3-4B-Instruct (left) and Qwen3-4B-Thinking (right).}
\label{fig:app_config_pair}
\end{figure*}

Figure~\ref{fig:app_config_pair} compares Qwen3-4B-Instruct and Qwen3-4B-Thinking on the same student trace. The example was randomly sampled from the 160 students evaluated by both configurations. The instruct model describes ``several explorations \ldots ranging from 0.5cm to 2cm,'' while the thinking model identifies two adjustments within 0.88--1.2 cm. The instruct output also attributes changes to ``system updates,'' whereas the thinking output identifies that the other variables remained constant. The two configurations consequently reach different adherence judgments. Across this example, 14 criterion ratings differ between the configurations, with 11 favoring the thinking model. The example illustrates the type of evidence-identification difference reflected in the aggregate RQ2 results.

\subsection{Effect of action-count guidance on an interpretation (RQ3)}
\label{app:ex_guidance}

\begin{figure*}[tp]
\centering
\includegraphics[width=\textwidth]{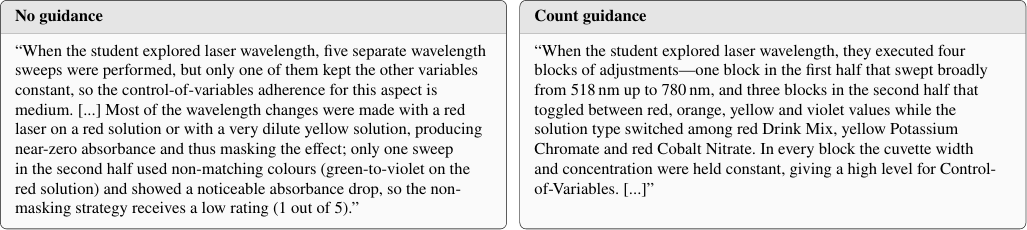}
\caption{The same \bl{} Laser Wavelength Optimization aspect interpreted by GPT-OSS-120B Thinking with \cop{}, without guidance (left) and with action-count guidance (right).}
\label{fig:app_guidance_pair}
\end{figure*}

Figure~\ref{fig:app_guidance_pair} compares GPT-OSS-120B with and without action-count guidance on the same student trace. The example was randomly sampled from the 160 students evaluated under both conditions. Without guidance, the model reports five sweeps, whereas the recorded count is four and assigns medium adherence. With the action count supplied, it reports four sweeps and assigns high adherence. The other two aspects in this example were counted correctly in both conditions.

\FloatBarrier

\end{appendices}

\end{document}